\documentclass[lettersize,journal]{IEEEtran}
\usepackage{amsmath,amsfonts}
\usepackage{algorithmic}
\usepackage{algorithm}
\usepackage{array}
\usepackage[caption=false,font=normalsize,labelfont=sf,textfont=sf]{subfig}
\usepackage{textcomp}
\usepackage{url}
\usepackage{verbatim}
\usepackage{graphicx}
\usepackage{cite}
\usepackage{booktabs}
\usepackage{multirow}
\usepackage{graphicx}
\usepackage{makecell}
\usepackage{multirow}
\usepackage{graphicx}
\usepackage{enumitem}
\usepackage{amsmath}
\usepackage{amssymb}
\usepackage{booktabs}
\usepackage{bm}
\usepackage{multirow}
\usepackage{multicol}
\usepackage{hhline}

\usepackage[dvipsnames]{xcolor}
\usepackage{colortbl} 
\definecolor{Gray}{gray}{0.05}
\usepackage{cancel}
\usepackage{comment}
\usepackage{cuted}
\usepackage{colortbl}
\usepackage{array}
\usepackage{amsmath,amsfonts}
\usepackage{pifont}
\usepackage[hidelinks]{hyperref}
\usepackage{threeparttable}

\begin{document}

\title{D-CLOT: Double Closed Loop Optimal Transport for Unsupervised Action Segmentation}
\author{Elena Bueno-Benito, Mariella Dimiccoli
\thanks{The authors are with the Institut de Rob\`otica i Inform\`atica Industrial, CSIC-UPC, Barcelona, Spain (e-mail: \{ebueno, mdimiccoli\}@iri.upc.edu).}}


\markboth{IEEE Transactions on Pattern Analysis and Machine Intelligence}%
{Bueno-Benito \MakeLowercase{\textit{and}} Dimiccoli: D-CLOT: Double Closed Loop Optimal Transport for Unsupervised Action Segmentation}

\maketitle

\begin{abstract}

Optimal transport (OT) has emerged as an effective framework for unsupervised action segmentation. Yet, in existing OT-based methods, the latent action prototypes that define the OT costs are not re-estimated from the refined frame geometry. Instead, they evolve solely through gradients from the pseudo-label loss. We identify this \emph{representation--prototype inconsistency} as a central bottleneck, particularly around ambiguous transitions and for short or infrequent actions. To address this issue, we build on the recently introduced CLOT, which refines frame embeddings based on estimated segment embeddings, and further re-estimates the action prototypes from the refined frame embeddings. Specifically, we introduce a graph-constrained module that regularizes the OT-refined frame and segment representations by preserving the local neighborhood geometry of the encoder output. An action-embedding refinement step then periodically re-anchors the prototypes to this stabilized representation geometry. We study two instantiations that share the same backbone, graph module, and objective: D-CLOT updates the prototypes using $k$-means, whereas D-CLOT$_{B}$ updates them as OT barycenters weighted by the refined transport plan, yielding an assignment-aware prototype update consistent with the current transport geometry. Across five established benchmarks, both variants improve segment-level quality over CLOT, with per-video gains of up to $+12.7$ F1 and $+10.2$ mIoU (YTI) and activity-level gains of up to $+8.9$ F1 (FS-Eval). We further establish the first unsupervised action-segmentation baseline on Assembly101, a procedural and substantially more fine-grained benchmark than those commonly used in prior work. Extensive ablations and sensitivity analyses demonstrate that the two refinement mechanisms are complementary and robust.
\footnote{Code, checkpoints, and V-JEPA2 features for Breakfast and Assembly101 will be released.}

\end{abstract}


\section{Introduction}
Temporal action segmentation assigns an action label to each frame of an untrimmed video and underlies downstream tasks such as activity recognition and procedural video understanding~\cite{ding2023survey,Lu2025Multimodal,li2024mvbench}. Although fully supervised methods achieve the strongest performance, they require costly frame-level annotations~\cite{Gong22,Wang2026}, motivating weakly supervised and fully unsupervised alternatives~\cite{Zhao2025,Huang2026,Xu_Weak2024}. Unsupervised action segmentation aims to discover recurring actions and their temporal boundaries without manual labels. Classical approaches first learn frame representations and then group them into action segments through clustering or temporal modeling~\cite{Kukleva2019,VidalMata2021,Li2021}. 

 \begin{figure}[t]
    \vspace{-1em}
    \includegraphics[width=1.0\linewidth]{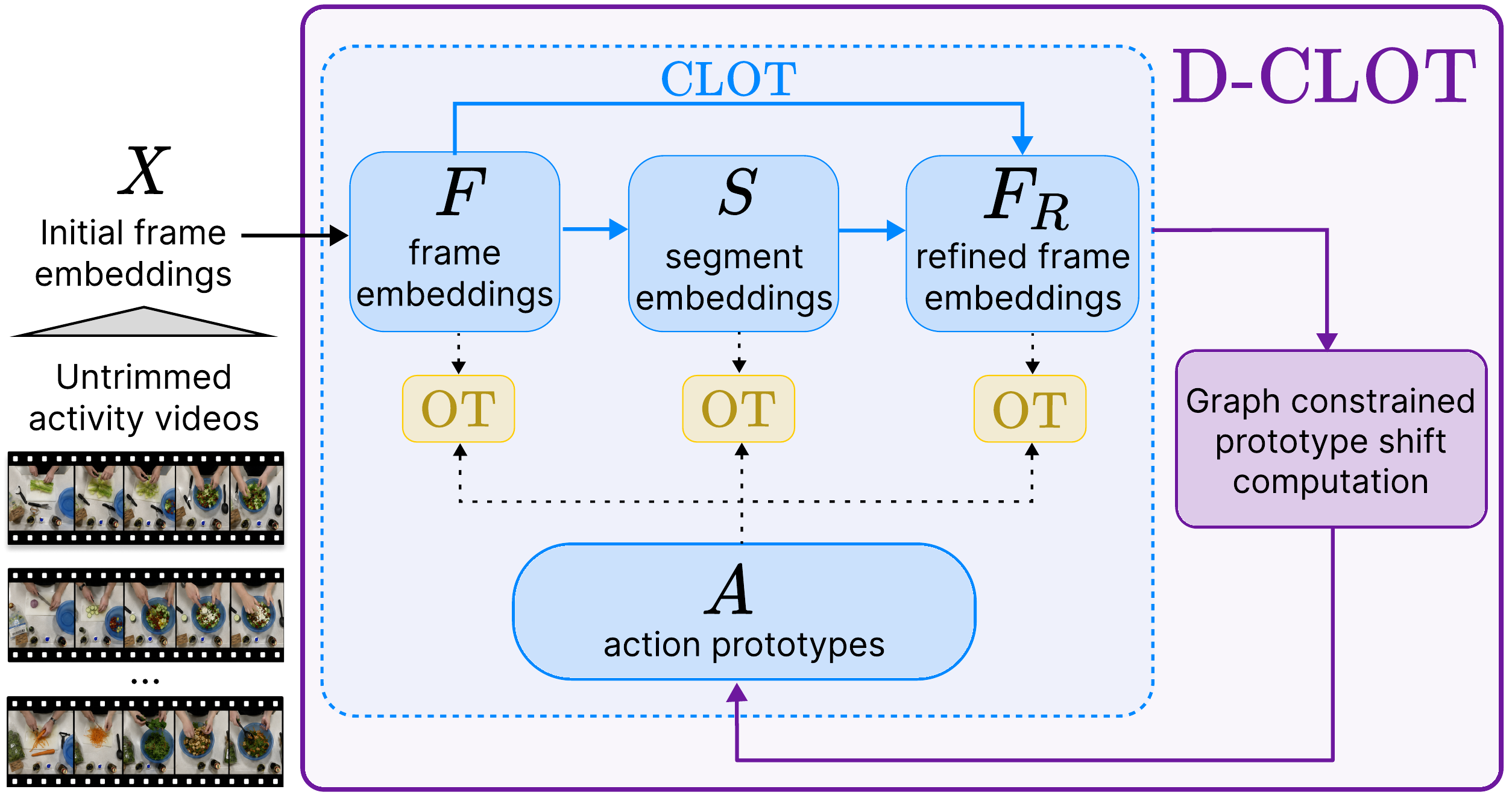}    
    \caption{\footnotesize Overview of the proposed D-CLOT framework. Given precomputed frame embeddings $\mathbf{X}$ from untrimmed videos, the CLOT loop refines frame- and segment-level representations through three OT-based alignment stages, producing frame embeddings $\mathbf{F}$, segment embeddings $\mathbf{S}$, refined frame embeddings $\mathbf{F}_R$, and their associated pseudo-labels. D-CLOT extends this refinement to the action prototypes $\mathbf{A}$, by computing their shift on a stabilized representation geometry, obtained through a graph regularization term. The blue boxes indicate learnable variables.}
    \vspace{-1.5em}
    \label{fig: CLOT2}
\end{figure}

A recent and effective line of work casts this joint problem as optimal transport (OT), aligning frame representations to latent action embeddings through a soft transport plan that doubles as a pseudo-label for self-training~\cite{Kumar22,Xu2024,ali2025}. ASOT~\cite{Xu2024} removed the need for a known action order by enforcing temporal consistency directly in the transport problem. HVQ~\cite{Spurio2024} improved short-action detection via hierarchical vector quantization. VASOT~\cite{ali2025} added cross-video correspondences as regularisation. CLOT~\cite{bueno-benito2025} unifies these threads through a multi-level cyclic mechanism: frame embeddings are aligned to action prototypes, a parallel decoder infers segment embeddings from the aligned sequence, and the resulting segment structure is propagated back to the frame domain via cross-attention before a final OT alignment.
 
This cyclic formulation leaves an asymmetry unaddressed: while frame and segment representations are iteratively refined, the action embeddings $\mathbf{A}$ that define the transport costs at every stage are not re-estimated from this refined geometry. They are initialised once via $k$-means and thereafter evolve only through the classification gradient, so as OT refinement re-organises the frame representation space, the prototypes meant to organise it can fall behind, a \emph{representation--prototype inconsistency}. This is most damaging around ambiguous temporal transitions, where refined embeddings of adjacent frames from different actions overlap, and for short or infrequent actions, whose limited transport mass is easily absorbed by dominant modes. Moreover, since the refinement stages optimise purely for the OT-induced classification loss, nothing prevents them from introducing associations between frames that were not neighbours in the original encoder geometry.

We address this problem with D-CLOT, which leaves the CLOT backbone and its three-stage OT objective intact, while introducing two successive operations. First, it stabilizes the refined frame and segment representations through a \emph{graph-constrained} regularizer that encourages their local neighbourhood structure to remain consistent with that of the encoder output. Second, it periodically re-anchors the action-embedding to this stabilized representation geometry, thereby mitigating the drift that can arise when they are updated solely through gradient descent. The proposed refinement is general and admits multiple prototype-update strategies. We study two instantiations that share the same backbone, graph module, and objective. D-CLOT refreshes $\mathbf{A}$ by applying $k$-means to the pooled stabilized representations, providing a lightweight, assignment-agnostic correction. In contrast, D-CLOT$_{B}$ updates each prototype as the OT barycenter of the stabilized representations, weighted by the soft mass of the refined transport plan, yielding an assignment-aware update consistent with the current transport geometry. Since the two variants differ only in their prototype re-estimation strategy, their comparison isolates the effect of assignment awareness on prototype refinement. We also broaden the empirical scope of the closed-loop OT paradigm 
as we move beyond the cooking and instructional domain with a new evaluation on Assembly101~\cite{assembly101}, a large-scale, fine-grained procedural-activity dataset. We define a controlled subset over 15 toy-assembly categories and two camera views, and extract V-JEPA\,2 features for it. With $11$--$42$ fine-grained actions per activity, far more than Breakfast (BF) \cite{breakfast}, YouTube Instructions (YTII)\cite{ytii}, 50Salads (FS) \cite{50salads}, or Desktop Assembly (DA)\cite{Kumar22}, this subset is a considerably harder testbed for unsupervised action segmentation.
 We evaluate D-CLOT on five datasets under activity- and video-level Hungarian matching, with an ablation isolating the graph module at the frame and refined-frame levels and the action-embedding refinement step, across both instantiations. 
The contributions of this paper are as follows:
\begin{itemize}
    \item We identify a representation-prototype inconsistency in CLOT: the latent action embeddings that define the OT costs are not re-estimated as the refined frame and segment representations evolve.
    \item We propose an action-embedding refinement step that periodically re-estimates the latent prototypes from the refined frame embeddings $\mathbf{F}^{R}$ and blends the resulting estimates with the current prototypes. Together with the graph constraint described below, this closes the representation--prototype loop without modifying the CLOT backbone or its training objective.
    \item We introduce a graph-constrained regularization module that stabilizes the OT-refined frame and segment representations by preserving the local neighborhood structure of the encoder's feature space.
    \item We study two instantiations of the proposed refinement: D-CLOT, which uses a $k$-means-based prototype update, and D-CLOT$_B$, which updates prototypes as OT barycenters weighted by the refined transport plan. Both variants achieve state-of-the-art performance on YTII~\cite{ytii}, FS~\cite{50salads}, DA ~\cite{Kumar22}, and BF ~\cite{breakfast}.
    \item We establish the first unsupervised action-segmentation baseline on Assembly101~\cite{assembly101}, a large-scale, procedural, fine-grained testbed.
\end{itemize}
The following section reviews related work. Section \ref{sec:methodology} introduces the proposed approach and details its two instantiations. Section \ref{sec:experiments} describes the validation protocol and discusses experimental results. Finally, Section \ref{sec:conclusions} concludes the paper.

\section{Related works}
\subsection{Temporal action segmentation}
 
Fully supervised action segmentation methods remain the most reliable but require costly data annotations \cite{huang2025, Bahrami2023, behrmann2022, Lu_2024}. To improve scalability and practicality, research has increasingly shifted towards weakly-supervised \cite{Lu2022, Souri2022, Xu_Weak2024, Zhang2023, Bueno-Benito2024} and unsupervised approaches \cite{Bueno-Benito2023, Ding2021, Kukleva2019, dias2018learning, Kumar22, Li2021, Li2024, Sarfraz2021,Tran23, VidalMata2021, Xu2024, Spurio2024, ali2025}, aiming to reduce reliance on labeled data while maintaining competitive segmentation performance.

As the estimated clusters lack semantic labels, the evaluation process requires finding the Hungarian correspondence between the clusters and the actual action classes. The Hungarian matching can be performed for video-level segmentation~\cite{Aakur2019,Bueno-Benito2023,Zexing2022,Li2024,Sarfraz2021}, activity-level segmentation~\cite{Ding2021,Kukleva2019,Kumar22,Li2021,Tran23,VidalMata2021,Xu2024,ali2025}, or for a global scope across an entire set of videos~\cite{Ding2021,Kukleva2019,Li2021, Bueno-Benito2024}. Depending on the hierarchical level used, methods aim to improve segmentation through these correspondences.

\noindent\textbf{Unsupervised video-level action segmentation} focuses on processing individual videos independently, without relying on predefined activity categories. Existing approaches can be broadly categorized into representation learning methods and clustering-based methods. Representation learning approaches aim to learn robust action features before applying a clustering algorithm. LSTM+AL~\cite{Aakur2019} predicts future frames and assigns segmentation boundaries based on prediction errors, while TSA~\cite{Bueno-Benito2023} introduces a contrastive learning framework that employs a triplet selection strategy. Clustering-based approaches directly segment videos using similarity metrics. While clustering has been underexplored in action segmentation, recent work on TW-FINCH~\cite{Sarfraz2021} incorporates temporal proximity alongside semantic similarity for improved clustering. Similarly, ABD~\cite{Zexing2022} detects action boundaries by measuring adjacent frame similarities, and OTAS~\cite{Li2024} enhances boundary detection by incorporating object-centric features.

\noindent\textbf{Unsupervised activity-level action segmentation} traditionally follows a two-step pipeline: first, learning action representations in a self-supervised manner, and then clustering the learned embeddings, typically assuming a predefined number of clusters. Classical methods strongly rely on temporal regularization to model the sequential nature of activities~\cite{Kukleva2019}. This idea has been further refined through encoder-decoder architectures, incorporating either visual reconstruction losses (VTE)~\cite{VidalMata2021} or discriminative embedding losses (UDE)~\cite{Swetha2021} to enhance clustering performance. Other methods have framed the problem as a self-supervised learning task, where action prototypes are discovered via auxiliary classification objectives~\cite{Ding2021, Li2021}. For instance, CAD~\cite{Ding2021} introduced a framework that identifies action prototypes using an activity classification task, while ASAL~\cite{Li2021} proposed a method that distinguishes between valid and invalid action orderings based on shuffled segment predictions.

Recently, OT has emerged as a powerful tool for jointly learning action representations and pseudo-labels through self-training, enabling effective feedback between representation learning and clustering, while directly optimizing for action segmentation. TOT~\cite{Kumar22} introduced a temporal OT formulation that generates pseudo-labels from predicted cluster assignments. However, this approach assumes fixed action ordering across all videos and uniform label assignment, contradicting the natural variability and long-tailed distribution of action labels. UFSA~\cite{Tran23} addressed these limitations by incorporating frame- and segment-level cues from transcripts, allowing for action permutations within activities and non-uniform label assignments. However, it still requires prior knowledge of an estimated action order to infer the segmentation.

To overcome this constraint, ASOT~\cite{Xu2024} introduced an OT-based method capable of producing temporally consistent segmentations without any prior assumptions about action order. This makes it particularly suited for both pseudo-labeling and decoding. However, ASOT enforces a strong structural prior, limiting its ability to detect short-duration actions, which are critical in many real-world applications. More recently, HVQ~\cite{Spurio2024} tackled this issue by using a hierarchical vector quantization-based approach, significantly improving short-action detection. However, its learned codebook representing action classes has limited generalization capabilities compared to ASOT, since it lacks explicit feedback between the representation and clustering. In a related direction, VASOT~\cite{ali2025} extends the OT-based paradigm by jointly addressing video alignment and action segmentation, leveraging cross-video correspondences to regularize the learned action representations. To jointly address ASOT's structural rigidity and HVQ's lack of representation-clustering feedback, CLOT~\cite{bueno-benito2025} introduced a closed-loop OT framework with a multi-level cyclic feature learning mechanism, in which frame and segment embeddings are iteratively refined through cross-attention and reconciled through three coupled OT problems.

\subsection{Optimal transport for structured prediction.}
OT has become a key framework in machine learning for measuring distributional discrepancies, particularly in unsupervised clustering and representation learning \cite{Xu2024, Kumar22, Tran23, bueno-benito2025, ali2025}. Unlike traditional probability metrics, Wasserstein distance preserves the geometric structure of distributions, making it well-suited for structured prediction tasks \cite{rabin2012wasserstein, lee2018wasserstein}. However, its computational complexity limits its scalability. To address this, projection-based OT (POT) methods, such as Sliced Wasserstein (SW) distance, efficiently approximate OT by projecting high-dimensional distributions onto lower-dimensional subspaces. SW distance has been successfully applied to point-cloud processing, color transfer, Gaussian Mixture Model learning, and domain adaptation, making it a practical alternative for large-scale applications \cite{kolouri2019generalized, nguyen2023sliced}.

 \subsection{Graph-constrained regularization and subspace clustering} Subspace clustering assumes that high-dimensional samples lie in a union of low-dimensional subspaces. Under the self-expressiveness principle, each sample is reconstructed from other samples, yielding a coding matrix $C$ with $X=XC$ and $\mathrm{diag}(C)=0$, from which an affinity matrix is built for spectral clustering \cite{vidal2011}. SSC \cite{elhamifar2013} promotes sparse affinities, LSR \cite{liu2012} enforces dense, grouping-consistent codings, and later work adds temporal structure, noise robustness, and outlier modelling \cite{li2019}. Such self-expressive models remain sensitive to the input representation: noisy or ambiguous features corrupt the learned affinity. Graph-constrained regularization mitigates this by letting an auxiliary representation adapt to the clustering objective while preserving the local neighbourhood geometry \cite{goyal2018}. In human motion segmentation: Dimiccoli et al.~\cite{dimiccoli2021} anchor jointly learned features and affinities to the original neighbourhood structure, while TVSH~\cite{xing2026} imposes a graph total-variation penalty that keeps the embedding smooth within segments yet sharp at boundaries and refines it from the segmentation output. These methods refine a static representation and derive the clustering from it. We differ in two respects: 1) we graph-regularize OT-refined frame features to obtain reliable geometry for updating latent action prototypes, and 2) we close the representation--prototype loop, maintaining their mutual consistency throughout refinement.
 
\subsection{Prototype and barycentric refinement} Prototype refinement is widely used in unsupervised representation learning to reduce representation drift and improve cluster consistency. Methods such as DeepCluster \cite{caron2018deepcluster} periodically update prototypes with $k$-means, while SwAV \cite{caron2020swav} integrates prototype assignments into online training through balanced clustering constraints. However, these methods are not designed for temporal action segmentation, where prototypes define the assignment geometry across temporally structured and duration-imbalanced actions. D-CLOT updates $A$ using $k$-means centroids, providing a simple prototype correction aligned with the current representation space. D-CLOT$_{B}$ instead updates $A$ as barycenters of the refined features weighted by the OT transport plan \cite{ cuturi2014fast, benamou2015iterative}. This barycentric update is assignment-aware, since each frame contributes according to its soft OT mass, making the refined prototypes consistent with the current transport geometry.

\begin{figure*}[t]
    \centering   
    \includegraphics[width=0.9\linewidth]{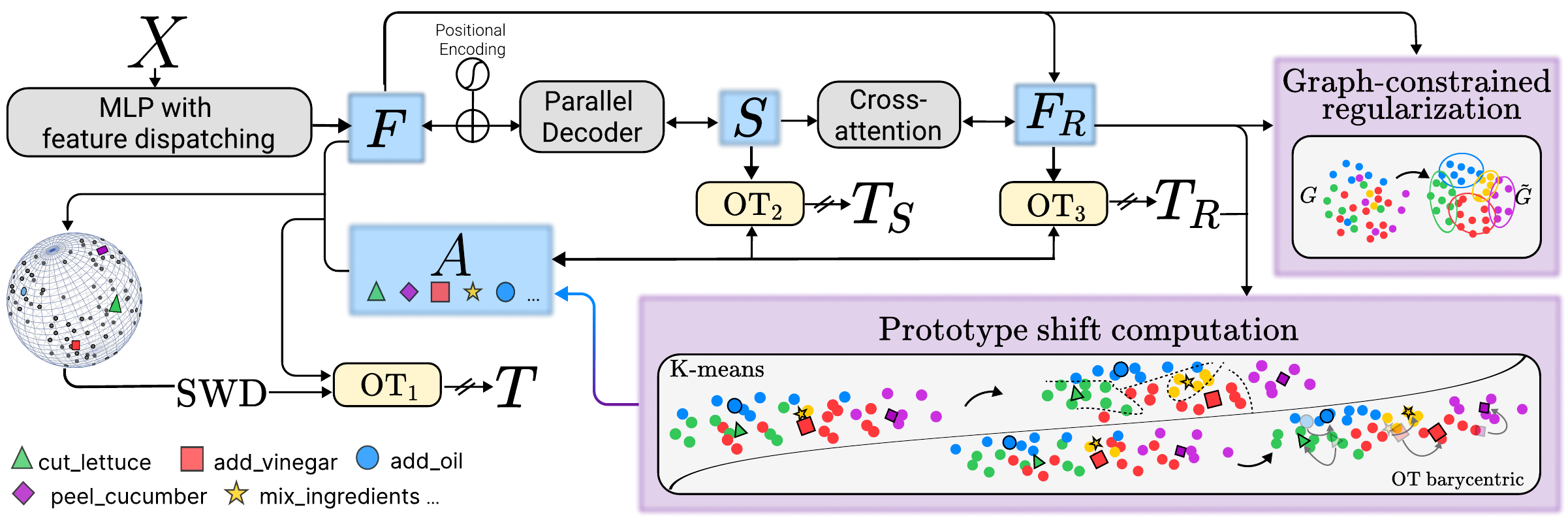}    
     
    \caption{\footnotesize Diagram of the \textbf{D-CLOT architecture}. 
    As in CLOT, an MLP encoder with feature dispatching produces frame embeddings $\mathbf{F}$, a parallel decoder estimates segment embeddings $\mathbf{S}$, and a cross-attention module propagates segment-level information back to the frame domain, yielding refined frame embeddings $\mathbf{F}_{R}$. Three OT problems are solved at the frame, segment, and refined-frame levels, producing the pseudo-label transport plans $\mathbf{T}$, $\mathbf{T}_{S}$, and $\mathbf{T}_{R}$, respectively. D-CLOT further introduces a graph-constrained module that regularizes the refined representation by preserving the local neighborhood structure of the encoded feature space.
     Then the regularized representation is used to explicitly update the action embeddings $\hat{\mathbf{A}}$ through an action-prototype refinement step. Two update variants are considered: a K-means refresh on $\mathbf{F}_{R}$ and an OT barycentric update using both $\widetilde{\mathbf{F}}_{R}$ and the refined transport plan $\widetilde{\mathbf{T}}_{R}$. The resulting update closes the representation--prototype loop by feeding the refined action embeddings back into the OT-based CLOT pipeline.}
    \label{fig:clotpp_architecture}
    \vspace{-1.5em}
\end{figure*}

\section{Methodology}
\label{sec:methodology}
We introduce D-CLOT, illustrated in Fig~\ref{fig:clotpp_architecture}, a graph-constrained action-refinement framework for unsupervised temporal action segmentation. Given a set of untrimmed videos, the goal is to infer frame-level action assignments without using frame-level labels, transcripts, or action-order constraints during training. The proposed method builds upon CLOT~\cite{bueno-benito2025}, which establishes a cyclic interaction between frame-level and segment-level representations through three optimal transport (OT) problems, and extends it by regularizing the refined frame geometry and explicitly updating the latent action embeddings from this refined representation.

Sections~\ref{subsec}~and~\ref{sec:ot} recall the CLOT backbone and OT formulation for completeness; Sections~\ref{sec:graph}~and~\ref{sec:action_refine} introduce the graph-constrained regularization and action-embedding refinement modules, which constitute the contributions of this extension.

\subsection{Problem Formulation}
Let $\mathcal{V} := \{V^b\}_{b=1}^{B}$ be a collection of $B$ untrimmed activity videos. Each video $V^b$ is a sequence of precomputed frame descriptors $\mathbf{X}^b = [\mathbf{x}_1^b;\ldots;\mathbf{x}_{N_b}^b] \in \mathbb{R}^{N_b \times D}$, where $N_b$ is the number of sampled frames and $D$ the input feature dimension. No frame-level labels, transcripts, or action-order constraints are available during training. Given a predefined number of latent action classes $K$, the objective is to infer $\hat{\mathbf{y}}^{b} = [\hat{y}^{b}_{1},\ldots,\hat{y}^{b}_{N_b}]$, $\hat{y}^{b}_{i}\in\{1,\ldots,K\}$, for every video $b$. The latent action space is parameterised by a shared action-embedding matrix $\mathbf{A} = [\mathbf{a}_{1},\ldots,\mathbf{a}_{K}]^{\top} \in\mathbb{R}^{K\times d}$, where $\mathbf{a}_{k}\in\mathbb{R}^{d}$ is the prototype of the $k$-th latent action, constrained to the unit sphere ($\|\mathbf{a}_k\|_2=1$). Importantly, $\mathbf{A}$ is shared across all videos of an activity and must therefore capture action structure consistently across subjects, durations, and orderings. For readability, we omit the video superscript $b$ whenever unambiguous. 

\subsection{Backbone}
\label{subsec}
 
D-CLOT retains the CLOT architecture as its backbone, mapping frame descriptors $\mathbf{X} \in \mathbb{R}^{N \times D}$ into three progressively structured representations: frame embeddings $\mathbf{F}$, segment embeddings $\mathbf{S}$, and refined frame embeddings $\mathbf{F}_{R}$. It comprises an MLP encoder with feature dispatching, a query-based parallel segment decoder, and a cross-attention refinement module.\vspace{1em}
 
\noindent\textbf{Encoder with feature dispatching.}
The encoder maps $\mathbf{X}$ into frame embeddings $\mathbf{F} = \mathrm{norm}(\mathrm{MLP}(\mathbf{X})) \in \mathbb{R}^{N \times d}$, where $d$ is the latent dimension and $\mathrm{norm}(\cdot)$ denotes $\ell_2$ normalisation. A feature-dispatching mechanism then injects information from the latent prototypes $\mathbf{A}$, conditioning each frame on its affinity to the prototype space. For each frame embedding $\mathbf{f}_{i}$ and prototype embedding $\mathbf{a}_{k}$, the affinity is
\begin{equation}
\phi(\mathbf{a}_{k}, \mathbf{f}_{i})
=
\sigma\!\left(
b_\phi+ s_\phi\,\frac{\mathbf{a}_{k}^{\top}\mathbf{f}_{i}}{\|\mathbf{a}_{k}\|_{2}\,\|\mathbf{f}_{i}\|_{2}}
\right),
\end{equation}
where $\sigma(\cdot)$ is the sigmoid and $b_\phi, s_\phi$ are learnable scalars, and the embedding is updated through a prototype-conditioned residual
\begin{equation}
\mathbf{f}'_{i} = \mathbf{f}_{i} + \frac{1}{K}\sum_{k=1}^{K}\phi(\mathbf{a}_{k}, \mathbf{f}_{i})\,\mathbf{a}_{k}.
\end{equation}

Each frame thus aggregates information from semantically related prototypes while preserving its descriptor through the residual. We write $\mathbf{F} = \mathrm{norm}(\{\mathbf{f}'_{i}\}_{i=1}^{N})$ for the dispatched representation. 
The resulting frame embeddings are aligned with the prototypes by a temporally consistent OT problem over a frame-to-cluster cost matrix $\mathbf{C}$, producing a soft frame-to-action assignment. 

\noindent\textbf{Parallel segment decoder.}
A transformer decoder maps $\mathbf{F}$ into segment embeddings $\mathbf{S} \in \mathbb{R}^{K' \times d}$. Rather than autoregressive decoding~\cite{behrmann2022}, it uses a parallel strategy adapted from action anticipation~\cite{Gong22}, inferring all $K'$ segment hypotheses jointly and avoiding the error accumulation of sequential decoders. A set of learnable queries $\mathbf{Q} \in \mathbb{R}^{K' \times d}$, each a latent segment prototype~\cite{Carion20}, interact through self-attention to model inter-segment dependencies and through cross-attention with the encoded frame sequence to gather frame-level evidence, with $K'\leq K$. As at the frame level, the decoder is coupled with a temporally consistent OT problem over a segment-to-cluster cost matrix $\mathbf{C}_{S}$.\vspace{1em}
 
\noindent\textbf{Cross-attention refinement.}
The final component propagates segment-level information back to the frame domain. Given the frame embeddings $\mathbf{F}$ and the segment embeddings $\mathbf{S}$, the refinement module computes a frame-to-segment attention map and updates the frame representation as
\begin{equation}
 \mathbf{F}_R = \mathbf{F} + \text{softmax}\!\left(\frac{\mathbf{F} \mathbf{S}^\top}{\tau \cdot\sqrt{d}}\right) \mathbf{S}
\end{equation}
where $\tau$ controls the sharpness of the alignment, so each frame aggregates its most relevant segment embeddings, combining local evidence with higher-level temporal context. The refined embeddings $\mathbf{F}_{R}$ are aligned with the prototype space through a third OT problem over a frame-to-segment-to-cluster cost matrix $\mathbf{C}_{R}$. The backbone thus forms a cyclic pipeline: frames are organised w.r.t.\ prototypes, segments are decoded from the frame sequence, and segment-level structure is projected back onto the frames. D-CLOT retains this pipeline and extends it as described next.

\subsection{Graph-Constrained representation regularization}
\label{sec:graph}
Although $\mathbf{F}_{R}$ incorporates segment-level information via cross-attention, the refined geometry may still inherit noise from the initial feature space, particularly near action boundaries and for rare or short-duration actions. Since OT pseudo-labels are sensitive to local feature geometry, when embeddings of temporally adjacent frames from different actions overlap, the transport plan assigns inconsistent soft labels that propagate errors into subsequent iterations. Therefore, we introduce a graph-constrained regulariser that encourages an auxiliary representation to preserve the local neighbourhood structure of the encoder anchor $\mathbf{F}_{\mathrm{enc}}$, while still allowing the CLOT objective to reorganise the space according to the latent action structure.\vspace{1em}

\noindent\textbf{Graph affinity.}
Given a representation $\mathbf{Z} \in \mathbb{R}^{N \times d}$ and a validity mask $\mathbf{m} \in \{0,1\}^N$ that zeros out padded frames, we define the normalised pairwise affinity $\mathbf{G} = \mathcal{S}(\mathbf{Z}) \in \mathbb{R}^{N \times N}_+$ as
\begin{align}
    G_{ij} = \frac{w_{ij}}{\displaystyle\sum_{(p,q)\in\mathcal{E}} w_{pq}},\,\;\;\;
  w_{ij} = m_i\, m_j\,
    \exp\!\left(-\frac{1 - \cos(\mathbf{z}_i,\mathbf{z}_j)}{h}\right),
    \label{eq:affinity}
\end{align}
where $\cos(\cdot,\cdot)$ is the cosine similarity on $\ell_2$-normalised rows, $\mathcal{E}$ is the set of valid frame pairs, and $h>0$ is a bandwidth. Diagonal entries are zero; optionally, a top-$k$ sparsification retains the $k$ largest weights per row, followed by symmetrisation. The matrix is finally normalised to unit total mass, so $\mathbf{G}$ encodes the local neighbourhood geometry of $\mathbf{Z}$ as an empirical joint distribution over frame pairs~\cite{goyal2018}.\vspace{1em}

\noindent\textbf{Graph-constrained loss.}
Let $\mathbf{F}_{\mathrm{enc}}$, the pre-dispatching encoder output, be a fixed anchor capturing the geometry of the input space with $\mathbf{G}^{0} = \mathcal{S}(\mathbf{F}_{\mathrm{enc}})$ be the affinity of the fixed anchor, and and $\mathbf{G} = \mathcal{S}(\mathbf{Z})$ that of an auxiliary representation $\mathbf{Z} \in \{\mathbf{F},\, \mathbf{F}_R\}$. The regulariser minimises the cross-entropy between the two induced affinity distributions,
\begin{equation}
  \mathcal{L}_\text{G}(\mathbf{F}_{\mathrm{enc}},\mathbf{Z})
    = -\sum_{i,j} G^{0}_{ij}\,\log\!\big({G}_{ij} + \epsilon\big),
  \label{eq:graph_loss}
\end{equation}
where $\mathbf{G}^{0}$ and $\mathbf{G}$ are computed from the auxiliary representation via~\eqref{eq:affinity} and $\epsilon>0$ is a numerical stabiliser. The anchor $\mathbf{F}_{\mathrm{enc}}$ is detached from the computation graph, so gradients flow only through $\mathbf{Z}$. Minimising~\eqref{eq:graph_loss} discourages the OT refinement stages from forming inter-action connections absent in the encoder geometry; because cross-entropy is strictly convex in the affinity distribution, the fidelity term admits a unique minimiser.

By default, the fidelity term is imposed as a soft penalty added to the CLOT objective (Section~\ref{sec:training_objective}), leaving $\mathbf{F}$ and $\mathbf{F}_R$ as the representations passed to the OT solver.\vspace{1em}

\subsection{Action-embedding refinement}
\label{sec:action_refine}
 
The graph module stabilises the refined frame geometry, but the action embeddings $\mathbf{A}$ are still updated only through gradient descent on the CLOT objective. As the OT stages reorganise the feature space, $\mathbf{A}$ can fall out of alignment with the stabilised geometry it is meant to organise. We therefore introduce an epoch-level \emph{action-embedding refinement} that periodically re-anchors $\mathbf{A}$ to the graph-stabilised representation. This leaves the CLOT architecture and the per-iteration gradient updates untouched: it is an outer loop that proposes an updated prototype set and blends it into $\mathbf{A}$.
 
To preserve cluster identity and avoid abrupt shifts, both variants apply a conservative, normalised blend rather than a hard replacement. Given a proposal $\hat{\mathbf{A}}$, the update is
\begin{equation}
  \mathbf{A} \leftarrow \mathrm{norm}\!\big((1-\beta)\,\mathbf{A} + \beta\,\hat{\mathbf{A}}\big),
  \label{eq:blend}
\end{equation}
with blend factor $\beta\in(0,1]$ and row-wise $\ell_2$ normalisation. The two configurations differ only in how the proposal $\hat{\mathbf{A}}$ is formed, and correspondingly in how the graph constraint of Section~\ref{sec:graph} is imposed: D-CLOT uses an assignment-agnostic $k$-means proposal on penalty-regularised features, while D-CLOT$_{B}$ uses an assignment-aware barycentric proposal on an explicitly graph-stabilised representation.\vspace{1em}
 
\noindent\textbf{D-CLOT: $k$-means refinement.}
In the first variant, we define $\mathcal{F} = \bigl\{(\mathbf{f}_{R}^{b})_{i} : b=1,\dots,B,\; i=1,\dots,N_b\bigr\}$
as the refined frame features pooled over all videos of the activity. The video index $b$ reappears here by necessity: the prototypes $K$ are shared across the activity and are therefore re-estimated from the entire collection rather than a single video. The proposal is
\begin{equation}
  \widehat{\mathbf{A}} \;\leftarrow\; \mathrm{KMeans}\!\left(\mathcal{F},\,K\right).
  \label{eq:kmeans_update}
\end{equation}
blended into $\mathbf{A}$ via~\eqref{eq:blend}. This update is assignment-agnostic: it uses no transport plan and no assumption beyond $K$. Its stability therefore rests entirely on the graph penalty~\eqref{eq:graph_loss} having kept $\mathbf{F}_R$ close to the encoder geometry, a dependence confirmed by the ablation in Section~\ref{sec:ablation}.\vspace{1em}
 
\noindent\textbf{D-CLOT$_{B}$: barycentric refinement.}
The barycentric proposal requires, for each frame, both a stabilised feature and a matched soft assignment. Rather than reading the penalty-regularised $\mathbf{F}_R$, this configuration enforces the graph constraint~\eqref{eq:graph_loss} through an explicit auxiliary variable. We solve a short inner problem
\begin{equation}
  \widetilde{\mathbf{F}}_R = \arg\min_{\mathbf{V}}\; \lambda_G\,\mathcal{L}_\text{G}(\mathbf{F}_{\mathrm{enc}},\mathbf{V}) + \lambda_{\text{a}}\,\big\|\mathbf{V}-\mathbf{F}_R\big\|_{F}^{2},
  \label{eq:aux_inner}
\end{equation}
where $\|\cdot\|_F$ is the Frobenius norm, approximated by a few $\ell_2$-normalised projected-gradient steps and used \emph{in place of} $\mathbf{F}_R$ for decoding and OT. The two terms play opposing roles: the fidelity term pulls the neighbourhood geometry of $\mathbf{V}$ towards the encoder anchor, while the alignment term keeps $\mathbf{V}$ close to the CLOT refined representation, so that $\lambda_G/\lambda_{\text{a}}$ directly controls how much the stabilised representation is allowed to depart from the OT-driven one. During training, the inner steps are differentiably unrolled, so gradients reach the network; at inference, they run without backpropagating to the network parameters. This yields an explicit graph-stabilised representation $\widetilde{\mathbf{F}}_R$ and its refined transport plan $\widetilde{\mathbf{T}}_R$.
 
Each prototype is then updated as the OT barycenter of the stabilised features, weighted by their soft assignment mass. Collecting the stabilised features $(\widetilde{\mathbf{f}}_{R}^{b})_{i}$ and the assignment entries $(\widetilde{\mathbf{T}}_{R}^{b})_{i,k}$ over the epoch, we define the total soft mass of action $k$ as $\pi_k = \sum_{b}\sum_{i} (\widetilde{\mathbf{T}}_{R}^{b})_{i,k}$ and set
\begin{equation}
  \hat{\mathbf{A}}_k =
  \begin{cases}
  \mathrm{norm}\!\left(\dfrac{\sum_{b}\sum_{i} (\widetilde{\mathbf{T}}_{R}^{b})_{i,k}\,(\widetilde{\mathbf{f}}_{R}^{b})_{i}}{\pi_k}\right), & \pi_k >  \varepsilon,\\[1.4ex]
  \mathbf{a}_k, & \text{otherwise,}
  \end{cases}
  \label{eq:barycenter_update}
\end{equation}
before blending via~\eqref{eq:blend}, where $\varepsilon>0$ guards the denominator against actions that receive no transport mass in a given epoch, and it is fixed to $10^{-8}$. Unlike the $k$-means update, the barycentric update is assignment-aware: each frame contributes to a prototype in proportion to its OT confidence, so the prototype tracks the current transport geometry, and frames with low mass contribute negligibly. This is what protects the long tail. A rare action whose frames are assigned with low confidence exerts a correspondingly small pull on its prototype, so the prototype is neither dragged towards a dominant mode nor displaced by ambiguous boundary frames.  

Both variants share the same backbone, the same graph-fidelity term~\eqref{eq:graph_loss}, and the same blended, normalised update~\eqref{eq:blend}. They differ in how the constraint is applied and in the resulting prototype proposal: D-CLOT imposes the graph term as a penalty and refreshes $\mathbf{A}$ by $k$-means over the refined features, whereas D-CLOT$_{B}$ enforces it through an explicit re-optimisation and updates $\mathbf{A}$ by an assignment-aware barycenter of the resulting $\widetilde{\mathbf{F}}_R$. Comparing them therefore isolates the contribution of assignment-awareness while holding everything else fixed.
\subsection{Optimal transport formulation}
\label{sec:ot}
 
For completeness we recall the unbalanced OT formulation of ASOT~\cite{Xu2024}, which underlies each of the three OT problems ($\mathbf{T},\mathbf{T}_S,\mathbf{T}_R$) in the pipeline; we refer to~\cite{Xu2024,bueno-benito2025} for the full derivation.\vspace{1em}
 
\noindent\textbf{KOT.}
Given histograms $\mu = \tfrac{1}{N}\mathbf{1}_{N}$ and $\nu = \tfrac{1}{K}\mathbf{1}_{K}$ and a cost matrix $\mathbf{C}\in\mathbb{R}^{N\times K}_{+}$, the Kantorovich problem~\cite{thorpe2018} seeks the minimum-cost coupling
\begin{equation}
\mathbf{T}^{*} = \min_{\mathbf{T}\in\mathcal{T}}\ \mathcal{F}_{\text{KOT}}(\mathbf{C},\mathbf{T}),\qquad \mathcal{F}_{\text{KOT}}(\mathbf{C},\mathbf{T}) := \langle \mathbf{C}, \mathbf{T}\rangle,
\end{equation}
over the transportation polytope $\mathcal{T} := \{\mathbf{T}\in \mathbb{R}_{+}^{N\times K}: \mathbf{T}\mathbf{1}_{K} = \mu,\ \mathbf{T}^{\top}\mathbf{1}_{N} =\nu\}$. Here $\mathbf{T}$ is a soft assignment between frames and actions.\vspace{1em}
 
\noindent\textbf{GW.}
The Gromov-Wasserstein term encodes structural priors such as temporal consistency by comparing intra-domain costs $\mathbf{C}^{v}\in\mathbb{R}^{N\times N}$ (frames) and $\mathbf{C}^{a}\in\mathbb{R}^{K\times K}$ (actions):
\begin{equation}
\mathcal{F}_{\text{GW}}(\mathbf{C}^{v},\mathbf{C}^{a},\mathbf{T}):= \!\!\sum_{\substack{i,k\in[N]\\ j,l\in[K]}}\!\! L(C^{v}_{ik},C^{a}_{jl})\,T_{ij}\,T_{kl},
\end{equation}
with $L:\mathbb{R}\times\mathbb{R}\to\mathbb{R}$ penalising deviations between cost entries. The two terms are fused as $\mathcal{F}_{\text{FGW}} := \alpha\,\mathcal{F}_{\text{GW}} + (1-\alpha)\,\mathcal{F}_{\text{KOT}}$ for $\alpha\in[0,1]$.\vspace{1em}
 
\noindent\textbf{Unbalanced relaxation.}
The action (column) marginal is relaxed into a soft KL penalty, allowing the long-tailed action distributions typical of untrimmed video:
\begin{equation}
\min_{\mathbf{T}\in\mathcal{T}_{p}}\ \mathcal{F}_{\text{FGW}}(\mathbf{C},\mathbf{T}) + \lambda\, D_{\mathrm{KL}}\!\big(\mathbf{T}^{\top}\mathbf{1}_{N}\,\big\|\,\nu\big),
\end{equation}
where $\mathcal{T}_{p} := \{\mathbf{T}\in\mathbb{R}_+^{N\times K} : \mathbf{T}\mathbf{1}_K=\mu\}$ keeps only the frame marginal, and $\lambda>0$ controls how strongly the action marginal $\mathbf{T}^{\top}\mathbf{1}_{N}$ is pulled towards $\nu$: larger $\lambda$ favours uniform usage, smaller $\lambda$ permits long-tailed distributions.\vspace{1em}
 
\noindent\textbf{Cost matrices.}
The stage-wise costs use $\{\mathbf{C}^{sw},\mathbf{C}^{v},\mathbf{C}^{a}\}$ for the frame and refined-frame problems and $\{\mathbf{C}^{k},\mathbf{C}^{s},\mathbf{C}^{a}\}$ for the segment problem. Except for $\mathbf{C}^{sw}$, they follow~\cite{Xu2024}: $\mathbf{C}^{v},\mathbf{C}^{a}$ penalise assigning temporally adjacent frames ($|i-k|\leq N_r,\ i\neq k$) to different actions ($j\neq l$), with no penalty outside the radius $N_r$ or when adjacent frames share an action. The visual cost is
\begin{equation}
\mathbf{C}^{k}_{ij} = \frac{\mathbf{x}_{i}^{\top}\mathbf{a}_j}{\|\mathbf{x}_i\|\,\|\mathbf{a}_j\|} - \rho\, Z_{ij},\qquad Z_{ij} = \big|\tfrac{i}{N}-\tfrac{j}{K}\big|,
\end{equation}
where $Z$ is the temporal prior of~\cite{Kumar22,Xu2024} regularising the coupling towards a banded-diagonal structure, weighted by $\rho\geq0$.\vspace{1em}
 
\noindent\textbf{Sliced-Wasserstein cost.}
For $\mathbf{C}^{sw}$ we complement the cosine cost with the Sliced-Wasserstein (SW) distance~\cite{kolouri2019generalized,nguyen2023sliced}. The $p$-Wasserstein distance between measures $\mu,\nu$ is
\begin{equation}
W_p(\mu, \nu) = \Big( \inf_{\gamma \in \Pi(\mu, \nu)} \int_{\mathbb{R}^d \times \mathbb{R}^d} \ell^p(x,y)\, d\gamma(x,y) \Big)^{1/p},
\end{equation}
estimated empirically by projecting onto $M$ random directions $\{\theta_m\}_{m=1}^{M}$ on $\mathcal{S}^{d-1}$,
\begin{equation}
\text{SWD}_p(\mathbf{x}_i, \mathbf{a}_j) = \Big( \tfrac{1}{M} \sum_{m=1}^{M} \ell\big( R_{\theta_m \#} \mathbf{x}_i, R_{\theta_m \#} \mathbf{a}_j \big) \Big)^{1/p},
\end{equation}
with $\ell$ the quadratic loss. We use $p=1$, reducing the SW distance to $M$ one-dimensional OT problems with closed-form solutions~\cite{rabin2012wasserstein}, and set
\begin{equation}
\mathbf{C}_{ij}^{sw} = 1 + \text{SWD}(\mathbf{x}_i,\mathbf{a}_j) - \mathbf{C}^{k}_{ij}.
\end{equation}
The SW cost is used at the frame and refined-frame levels; the simplified variant omits it, using $\mathbf{C}^{k}$ alone.
 

\subsection{Training objective and inference}
\label{sec:training_objective}
 
We train in a fully unsupervised manner by minimising the cross-entropy between predicted assignment probabilities and OT pseudo-labels at all three levels. For video $b$, the frame-to-action probabilities are $P^b_{ij} = \mathrm{softmax}(\frac{\mathbf{F}^b \mathbf{A}^\top}{\tau})_{ij}$ with temperature $\tau>0$, and the masked cross-entropy is
\begin{equation}
\mathcal {L}(\mathbf{T}, \mathbf{P}) = -\frac {1}{B}\sum _{b=1}^B \sum_{i=1}^N\sum_{j=1}^K T^b_{ij} \log P^b_{ij},,
\end{equation}
The segment and refined-frame probabilities $\mathbf{P}_S,\mathbf{P}_R$ are defined analogously against $\mathbf{T}_S,\mathbf{T}_R$, giving the three-stage CLOT loss $\mathcal{L}_{\mathrm{CLOT}} = \mathcal{L}(\mathbf{T},\mathbf{P}) + \mathcal{L}(\mathbf{T}_S,\mathbf{P}_S) + \mathcal{L}(\mathbf{T}_R,\mathbf{P}_R)$.
 
The graph regulariser enters the objective only under the penalty mechanism (D-CLOT), giving the per-iteration loss
\begin{equation}
  \mathcal{L}^{\text{train}}_{D-CLOT} = \mathcal{L}_{\mathrm{CLOT}} + \lambda_{G}^{F}\,\mathcal{L}_{G}(\mathbf{F}_{\mathrm{enc}},\mathbf{F}) + \lambda_{G}^{R}\,\mathcal{L}_{G}(\mathbf{F}_{\mathrm{enc}},\mathbf{F}_R),
  \label{eq:clotpp_loss}
\end{equation}
with $\lambda_{G}^{F},\lambda_{G}^{R}\geq0$ weighting the frame- and refined-frame-level regularisation. Under the auxiliary-variable mechanism (D-CLOT$_{B}$) the fidelity term is instead consumed inside the inner problem~\eqref{eq:aux_inner}, and the outer objective reduces to $\mathcal{L}_{\mathrm{CLOT}}$ evaluated on the graph-stabilised representations, $\widetilde{\mathbf{F}}$ and $\widetilde{\mathbf{F}}_R$.
 
At inference, we obtain $\mathbf{F}_R$. Under the auxiliary-variable mechanism, this representation is graph-stabilised via the same inner steps used during training, without backpropagation. The final segmentation solves the refined OT problem between $\mathbf{F}_R$ (or $\widetilde{\mathbf{F}}_R$) and $\mathbf{A}$, assigning each frame $\hat{y}_i = \arg\max_{k}(\mathbf{T}_{R})_{i,k},$ with $\mathbf{T}_{R}$ replaced by $\widetilde{\mathbf{T}}_{R}$ for D-CLOT$_{B}$.

\section{Experiments}
\label{sec:experiments}

\subsection{Experimental setting}
\paragraph{Implementation details}
The backbone is unchanged from CLOT~\cite{bueno-benito2025} (Section~\ref{subsec}). Optimisation uses Adam with weight decay $10^{-4}$ and a learning rate around $10^{-3}$. Action embeddings $\mathbf{A}$ are initialised by $k$-means, and the number of action clusters $K$ is set to the ground-truth action count of each dataset, consistent with prior work~\cite{Spurio2024, Xu2024, ali2025, bueno-benito2025}. Videos are sampled to $256$ frames per training iteration, except on Assembly101~\cite{assembly101}, where the longer average duration motivates a budget of $512$ frames. Training alternates gradient-based optimisation of~\eqref{eq:clotpp_loss} with periodic re-estimation of $\mathbf{A}$: the prototypes are refreshed by $k$-means~\eqref{eq:kmeans_update} for D-CLOT or by the OT barycenter~\eqref{eq:barycenter_update} for D-CLOT$_{B}$, and blended into the current prototypes via~\eqref{eq:blend} rather than replaced outright. Neither module introduces learnable parameters, so both variants have the same parameter count as CLOT. Detailed hyperparameter settings are reported in supp.mat.
\paragraph{Datasets and Features.} 
We evaluate our approach on five video datasets covering cooking, instructional, and assembly-oriented procedural activities. 
Following standard practice in unsupervised action segmentation, we rely on pre-extracted visual representations to ensure a fair comparison with prior work~\cite{Tran23, Kukleva2019, Kumar22, Spurio2024, Xu2024, Ding2021}.  Table~\ref{tab:dataset-stats} reports the principal statistics of each dataset.

\begin{itemize}
    \item \textbf{Breakfast (BF)}~\cite{breakfast} contains approximately $1{,}700$ videos of subjects preparing breakfast dishes, organized into $10$ activities and $48$ fine-grained action classes (e.g., \emph{cracking an egg} or \emph{pouring flour}). Clip duration ranges from about $30$~s to several minutes. We use the standard Fisher-vector encoding of improved dense trajectories (IDT)~\cite{IDT}.

    \item \textbf{YouTube Instructions (YTI)}~\cite{ytii}comprises $150$ instructional
    videos from $5$ activities, each lasting roughly two minutes. It is particularly
    challenging because background frames dominate the footage (approximately $75\%$).
    We represent each frame by the concatenation of HOF descriptors and VGG16 conv5
    features~\cite{VGG}.

\begin{table}[t]
    \centering
    \caption{Statistics of datasets used in the experiments. Background means the \% of background frames in a dataset.}
    \label{tab:dataset-stats}
    \resizebox{\columnwidth}{!}{
    \begin{tabular}{lrrrrrr}
        \toprule
        \textbf{Statistic} & \textbf{BF} & \textbf{YTII} & \textbf{FS (mid)} & \textbf{FS (eval)} & \textbf{DA} & \textbf{A101} \\
        \midrule
        \#Videos                                       & 1{,}712   & 149       & 50        & 50        & 76        & 200         \\
        Avg.\ \#Frames/video                       & 2{,}097.49 & 516.60   & 11{,}551.90 & 11{,}551.90 & 778.49  & 13{,}047.81 \\
        \quad Minimum                                  & 130       & 95        & 7{,}555   & 7{,}555   & 611       & 4{,}429     \\
        \quad Maximum                                  & 9{,}741   & 2{,}320   & 18{,}143  & 18{,}143  & 1{,}154   & 55{,}248    \\
        Feature Dim.                                   & 64        & 3{,}000   & 64        & 64        & 512       & 1{,}408     \\
        \#Activities (V)                               & 10        & 5         & 1         & 1         & 1         & 15          \\
        Avg.\ \#Actions/video                & 5      & 9      & 18     & 17     & 22    & 15       \\
        
        Background                                     & 12.1\%   & 61.9\%   & 14.4\%   & 14.4\%   & 2.7\%    & 42.1\%     \\
        \bottomrule
    \end{tabular} }
    \vspace{-0.2in}
\end{table}
    \item \textbf{50 Salads (FS)}~\cite{50salads} includes $50$ salad-preparation videos totaling $4.5$ hours. We report results under the two standard annotation granularities: Mid, with $19$ action classes, and Eval, with $12$ coarser classes. We use Fisher vector features extracted from IDT~\cite{IDT}.

    \item \textbf{Desktop Assembly (DA)}~\cite{Kumar22} contains $76$ videos, each of
    about $1.5$~min, in which subjects assemble a desktop setup following a fixed
    procedure of $22$ temporally ordered actions. We use the features released by
    Tran et al.~\cite{Tran23}.

   \item \textbf{Assembly101}~\cite{assembly101}  is a large-scale multi-view dataset for toy assembly and disassembly. The full dataset comprises $4,321$ videos, with an average duration of approximately $7.1$ minutes and a total duration of about $513$ hours. It covers $15$ toy categories and provides annotations for $202$ coarse action classes. Since the full benchmark contains a large number of videos, viewpoints, and fine-grained action instances, we define a controlled subset tailored to unsupervised action segmentation. Specifically, we use the coarse action annotations, restrict the data to the \textit{assembly} task, and evaluate the two locally available views, \texttt{view1} and \texttt{view2}. The resulting subset contains $200$ view-specific videos from the local train--validation split and spans all $15$ toy categories. The number of observed coarse actions per toy category ranges from $11$ to $42$, from \textit{jackhammer} to \textit{excavator}. We extract V-JEPA2 features~\cite{assran2025vjepa2} and treat each view-specific recording as an independent video instance\footnote{The pretrained weights used in this work are available at \href{https://huggingface.co/facebook/vjepa2-vitl-fpc64-256}{\nolinkurl{huggingface.co/facebook/vjepa2-vitl-fpc64-256}}.}. 
\end{itemize}

 \begin{table*}[ht!]
 \caption{Comparisons of action segmentation performance obtained by applying the \textcolor{blue}{Hungarian matching} per \textcolor{blue}{video} and at the \textcolor{blue}{activity-level} "Full" on the Breakfast~\cite{breakfast}, Youtube Instr. ~\cite{ytii},  50Salads ~\cite{50salads} and Desktop Assembly \cite{Kumar22} benchmarks. The highest accuracy is in \textbf{bold}, and the second highest is \underline{underlined}. \textcolor{NavyBlue}{\texttt{D-CLOT}} refreshes action embeddings with K-means on refined frame embeddings, while \textcolor{Blue}{\texttt{D-CLOT$_{B}$}} updates action embeddings as barycenters of graph-constrained refined features. }
    \centering
    \setlength\tabcolsep{3pt}
    \resizebox{6.0in}{!}{
    \begin{tabular}{@{}llccccccccccccccc@{}}
    \toprule
      & \multirow{2}{*}{Methods} & \multicolumn{3}{c}{Breakfast} &  \multicolumn{3}{c}{YTI} &\multicolumn{3}{c}{50Salads (Mid)} & \multicolumn{3}{c}{50Salads (Eval)} & \multicolumn{3}{c}{DA} \\
    \cmidrule(lr){3-5} \cmidrule(lr){6-8} \cmidrule(lr){9-11} \cmidrule(lr){12-14} \cmidrule(lr){15-17}
   & & MoF & F1 & mIoU & MoF & F1 & mIoU & MoF & F1 & mIoU & MoF & F1 & mIoU & MoF & F1 & mIoU  \\
    \midrule
  Per video &  TWF*~\cite{Sarfraz2021} & 62.7  & 49.8  & 42.3  & 56.7  & 48.2  & - & 66.8  & 56.4  & \underline{48.7}   &\underline{71.7} &-  & -  &73.3  & 67.7  & \textbf{57.7}\\
   & ABD*~\cite{Zexing2022} & 64.0 & 52.3 & - & 67.2 & 49.2 & - & \underline{71.8} & - & - & 71.2 & - & - & - & - & - \\
   & OTAS*~\cite{Li2024} &\textbf{67.9} & - & - & 65.7 & - & - & \textbf{72.4}& - & - & \textbf{73.5} & - & - & - & - & - \\
  &  TSA*~\cite{Bueno-Benito2023} \small(kmeans) & 63.7 & 58.0 & \textbf{53.3} & 59.7& 55.3  & - &-&-&-&-&-&-&-&-&-\\
    
   \arrayrulecolor{gray!25}
    \cmidrule(lr){2-16}
    \arrayrulecolor{black}
    
  &   ASOT~\cite{Xu2024} & 63.3 & 53.5 & 35.9 & 71.2 & 63.3 & 47.8 & 64.3 & 51.1 & 33.4 & 64.5 & 58.9 & 33.0 &73.4 & 68.0 & 47.6\\

   &  CLOT~\cite{bueno-benito2025} & 66.3 & 55.9 & 37.1& 69.3 & 60.8 & 48.2 &  69.4 & 63.8  & 45.0 & 64.6 & 69.7 & 42.5 &  73.5& 75.2  & 52.4  \\    

  \rowcolor{NavyBlue!6}
\cellcolor{white} & D-CLOT  &65.7& \underline{58.1} & \underline{38.6}  &  \textbf{74.9}  &  \textbf{73.5} & \textbf{58.4} &  70.5 & \textbf{70.0} & \textbf{49.2} & 67.9 & \underline{71.9} & \underline{43.7} & \underline{73.7} & \underline{79.0}  & 54.9   \\
  \rowcolor{NavyBlue!13}
\cellcolor{white} &  D-CLOT$_{B}$  &   \underline{67.5} & \textbf{59.0} & \underline{38.6} & \underline{71.6}  & \underline{65.9}   & \underline{49.7} &  69.4 & \underline{66.2} & 46.1 & 68.7 & \textbf{75.6} & \textbf{44.7} & \textbf{74.5} & \textbf{79.2}& \underline{55.2} \\
    \midrule

   Full  &CTE~\cite{Kukleva2019} &  41.8 & 26.4 & - & 39.0 & 28.3  & - &30.2   & -  & -  & 35.5  & - & - &47.6 & 44.9& -\\
  & VTE~\cite{VidalMata2021} & 48.1  & - & - & - & 29.9  &-  & 24.2  & -  & -  & 30.6  & - & - &-  &-  &- \\
 &   UDE~\cite{Swetha2021} & 47.4  & 31.9 & - & 43.8 & 29.6  & - & -  &  - & - &  42.2 & 34.4  & - & - & - & - \\

  &  ASAL~\cite{Li2021} & 52.5  & 37.9 & - & 44.9 & 32.1  & - & 34.4  & -  & -  & 39.2  &-  &-  &-  &-  &- \\
    
   & TOT~\cite{Kumar22}  &47.5 & 31.0 & - &40.6 & 30.0 & - &31.8 & - & - &47.4 & 42.8 & - &56.3 & 51.7 & -   \\
   & TOT+~\cite{Kumar22} &39.0 & 30.3 & - &45.3 & 32.9 & - &34.3 & - & -& 44.5 & 48.2 & - &58.1 & 53.4 & - \\
   & UFSA~\cite{Tran23} &52.1 & 38.0 & - & 49.6 & 32.4 & - & 36.7 & 30.4 & - & 55.8 & 50.3 & -& 65.4 & 63.0 & - \\
   &  ASOT~\cite{Xu2024} &  56.1 & 38.3 & 18.6 & 52.9 & 35.1 & 24.7 & 46.2 & 37.4 & 24.9 & 59.3 & 53.6 & 30.1 & 70.4 & 68.0 & 45.9\\
  &  HVQ~\cite{Spurio2024} & 54.4  & 39.7 & - & 50.3 & 35.1  & - & -  & -  & -  &-   & - & - & - & - & - \\
    
   &  VASOT~\cite{ali2025} &57.5 & 39.0 & 18.8&  53.2& 35.7& 25.2& 47.2 & 41.3& 26.1&  60.6  & 57.4 & 34.5 &\textbf{70.9}  &  75.1  & 49.3\\
   &  CLOT~\cite{bueno-benito2025} & 60.1  & 40.1 &  18.5& 54.4 & 36.7  & 23.4 & 50.6  & 46.6 & 31.4 &   59.4 & 63.2 & 38.8   & 68.8 & 72.6 & 48.1 \\
    
    
  \rowcolor{NavyBlue!6}
\cellcolor{white} & D-CLOT  & \underline{59.4}& \underline{40.5} & \underline{19.0}  &  \textbf{55.7}  & \textbf{37.9}  & \textbf{27.3} & \textbf{51.0} & \textbf{47.8} & \textbf{32.8} & \underline{62.7} & \underline{65.3} & \underline{39.1} & 68.7 & \textbf{76.4}  & \underline{50.5}   \\
  \rowcolor{NavyBlue!13}
\cellcolor{white} &  D-CLOT$_{B}$  & 58.5  & \textbf{41.3}&\textbf{19.1} &\underline{54.9}  & \underline{36.8}  & \underline{26.4}  &  \underline{50.8} & \underline{47.0} & \underline{32.0} & \textbf{64.9} & \textbf{72.1} &  \textbf{40.4} &  \underline{69.7} & \underline{76.2}& \textbf{51.8} \\
    \bottomrule
    \end{tabular}}
    \vspace{-0.1in}
    
    \label{tab:sota_activity}

\end{table*}

\begin{table}[t]
\centering
\caption{
Unsupervised action segmentation results on Assembly101~\cite{assembly101}.
Results are reported using V-JEPA2 features~\cite{assran2025vjepa2} over 15 toy-category activities, with \textcolor{blue}{Hungarian matching} computed at the \textcolor{blue}{activity level} and per \textcolor{blue}{video}. }
\label{tab:assembly101_baseline}
\resizebox{0.95\columnwidth}{!}{
\begin{tabular}{l ccc ccc c}
\toprule

& \multicolumn{6}{c}{Assembly101} &
  \\
 \toprule

\multirow{2}{*}{Method} & \multicolumn{3}{c}{Video-level}
& \multicolumn{3}{c}{Activity-level} 
& \multirow{2}{*}{Avg.} \\
\cmidrule(lr){2-4} \cmidrule(lr){5-7}
  & MoF & F1 & mIoU & MoF & F1 & mIoU & \\
\midrule

CLOT~\cite{bueno-benito2025}  
& 53.1 & 42.2 & 21.8  & 26.1 & 11.5& 4.8 

& 26.6 \\

 \rowcolor{NavyBlue!6}D-CLOT 
& 49.5 & 54.7 & 37.9 & 24.8 & 15.3 & 10.3

& 32.1 \\

     \rowcolor{NavyBlue!13} D-CLOT$_{B}$  
& 49.0 & 56.5 & 40.4 & 24.7 & 16.2 & 9.5

& \textbf{32.7} \\

\bottomrule
\end{tabular}
}
\vspace{-1em}
\end{table}

\paragraph{Metrics.} 
We follow standard evaluation protocols for unsupervised action segmentation~\cite{ding2023survey}. Since predicted labels are unordered clusters, they are aligned with ground-truth action labels using Hungarian matching. We report two complementary settings: video-level matching, where the alignment is computed independently for each video~\cite{Sarfraz2021, Zexing2022, Bueno-Benito2023}, and activity-level matching, where a single alignment is computed over all videos of the same activity category~\cite{Tran23, Kukleva2019, Kumar22, VidalMata2021, Spurio2024, Xu2024}. For Assembly101~\cite{assembly101}, where the proposed benchmark is defined over selected toy categories, we analogously report toy-level matching over the selected subset.
We report three standard metrics: \textbf{Mean over Frames (MoF)}, which measures frame-wise accuracy and is sensitive to dominant actions; \textbf{F1 Score}~\cite{Kukleva2019}, which evaluates segment-level agreement; and \textbf{mean Intersection over Union (mIoU)}, which averages class-wise overlap and provides a more balanced assessment under class imbalance. Since MoF is less sensitive to short and rare actions, F1 and mIoU are the primary metrics for evaluating this failure mode.

\subsection{State-of-the-art comparisons} 

Table~\ref{tab:sota_activity} compares D-CLOT and D-CLOT$_{B}$ against prior unsupervised methods under both per-video and activity-level Hungarian matching. Both differ only in how the action prototypes are re-estimated once the frame and segment representations have been refined: D-CLOT applies $k$-means to the stabilized embeddings (assignment-agnostic), whereas D-CLOT$_{B}$ updates each prototype as an OT barycenter weighted by the refined transport plan (assignment-aware).

\noindent\textbf{Per-video matching.} D-CLOT attains the best F1 and mIoU on YTI and FS-Mid, improving over CLOT by $+12.7$ F1 and $+10.2$ mIoU on YTI, and by $+6.2$ F1 and $+4.2$ mIoU on FS-Mid. D-CLOT$_{B}$ provides the strongest gains on FS-Eval and DA, increasing F1 by $+5.9$ and $+6.3$ points, respectively, and mIoU by $+2.2$ and $+4.3$ points. On YTI, both variants also improve MoF, so the gain there is not a trade-off but a uniform improvement. Both variants surpass methods that train directly on the target videos (TWF*, ABD*, OTAS*, TSA*) on most metrics, without requiring any per-video fine-tuning. 

\noindent\textbf{Full matching.} The same pattern holds under the harder activity-level protocol, and on F1 and mIoU both variants improve on all OT-based baselines across every dataset. D-CLOT leads on YTI ($+1.3$ MoF, $+1.2$ F1, $+3.9$ mIoU over CLOT) and FS-Mid ($+0.4$ MoF, $+1.2$ F1, $+1.4$ mIoU), while D-CLOT$_{B}$ is best on FS-Eval ($+5.5$ MoF, $+8.9$ F1, $+1.6$ mIoU) and attains the top DA mIoU, alongside D-CLOT's best DA F1. On Breakfast, both variants trade a small MoF decrease for higher segment-level scores: D-CLOT$_{B}$ improves F1 by $+1.2$ and mIoU by $+0.6$, while D-CLOT yields smaller gains ($+0.4$ F1, $+0.5$ mIoU) with a smaller MoF drop ($-0.7$). This trade-off is expected: MoF is dominated by long-duration actions, whereas F1 and mIoU reward class-balanced, segment-level agreement, precisely what prototype re-anchoring improves near ambiguous transitions and for short or infrequent actions.

Overall, the proposed variants achieve the best or second-best results on most datasets and metrics. The choice between variants is dataset-dependent: D-CLOT tends to benefit datasets with sharper cluster structure (YTI, FS-Mid), while the assignment-aware update of D-CLOT$_{B}$ is more effective on datasets with more procedurally consistent, imbalanced action durations (BF, FS-Eval, DA).

\begin{table*}[ht!]
    \centering
    \caption{\footnotesize \textbf{Ablation Study on the four datasets}: Breakfast~\cite{breakfast}, YTI~\cite{ytii}, 50Salads~\cite{50salads} and Desktop Assembly~\cite{Kumar22}. GC denotes the graph-constrained term, applied at the frame level ($F$), the refined-frame level ($F_R$), or both; $\hat{A}$ denotes the periodic re-estimation of the prototypes. The top row (CLOT~\cite{bueno-benito2025}) is the baseline. Within each variant block, the highlighted row is the full model (GC at both levels \emph{and} refinement); the other rows toggle one factor at a time. Rows labelled \textbf{"only GC in $\cdot$"} activate the graph term at the indicated level(s) \emph{without} refining the prototypes, thus isolating GC alone. Rows labelled \textbf{"w/o GC $\cdot$"} keep refinement active but remove the graph term at the indicated level. Both full models improve over CLOT on most metrics, and the full D-CLOT$_B$ is the only configuration that surpasses \emph{all} its ablated variants on every dataset and metric. }
    \setlength\tabcolsep{3pt}
    \resizebox{6.0in}{!}{
        \begin{tabular}{@{}ll*{15}{c}@{}}
        \toprule
         \multirow{2}{*}{} & & \multicolumn{3}{c}{Breakfast} &  \multicolumn{3}{c}{YTI} &\multicolumn{3}{c}{50Salads (Mid)} & \multicolumn{3}{c}{50Salads (Eval)} & \multicolumn{3}{c}{DA} \\
         \cmidrule(lr){3-5} \cmidrule(lr){6-8} \cmidrule(lr){9-11} \cmidrule(lr){12-14} \cmidrule(lr){15-17}
         && MoF & F1 & mIoU & MoF & F1 & mIoU & MoF & F1 & mIoU & MoF & F1 & mIoU & MoF & F1 & mIoU  \\
        \midrule
        CLOT~\cite{bueno-benito2025}   &  & 60.1  & 40.1 &  18.5 & 54.4& 36.7 & 23.4 & 50.6  & 46.6 & 31.4 & 59.4 & 63.2 & 38.8& 68.8 &72.6& 48.1 \\
        \arrayrulecolor{gray!25}
        \midrule
        \arrayrulecolor{black}
        \textbf{D-CLOT} w/o $\hat{A}$ & only GC in $F$  & 57.5  & 39.4  &17.9 & 51.8 & 36.0  &26.9   & 51.2  &46.3 &31.3 &49.5 & 57.1 & 36.4 &60.5 &67.5 &41.7\\
       shared by both variants     & only GC in $F_R$  & 58.3  &  40.2 &  18.3   & 50.8  & 34.8 & 25.8 & 50.6  &47.5 &32.4 &  48.3&56.4 &36.0  & 65.8& 68.8&45.0\\
        
         &only GC in $F,\,F_R$  & 58.1  & 39.6  & 18.1  & 50.1 & 34.7 & 25.1 &  51.9 & 45.3& 30.1 & 48.2  &56.2 & 36.1 &58.3 & 64.1&40.3\\
        \arrayrulecolor{gray!25}\midrule\arrayrulecolor{black}
        \textbf{D-CLOT} & w/o GC  & 58.3 & 39.8    & 17.9 & 53.9 & 36.2 & 25.4 &  46.3 &42.8 & 29.4 & 50.1 &58.1 &36.1  &58.8 &59.2 &38.7\\ 
        
         & w/o GC in $F_{R}$& 58.5 & 40.3 & 18.0 &54.6 & 36.8  & 26.2 & 49.7  & 47.5& 32.0&51.9  &59.4 & 37.3 & 59.2& 61.8&39.6\\
          &w/o GC in $F$ & 59.5  & 41.0   & 18.8&54.4 & 37.7  &26.3  &  50.6 & 47.0& 32.2&  54.9& 66.6 & 38.9 & 67.0& 68.9&46.7\\
      \rowcolor{NavyBlue!6}
    \cellcolor{white} & Full &59.4 & 40.5 & 19.0  &  55.7  & 37.9  & 27.3 & 51.0 & 47.8 & 32.8 & 62.7 &  65.3 & 39.1 & 68.7 & 76.4 & 50.5   \\
        
          \arrayrulecolor{gray!25}
        \midrule
        \arrayrulecolor{black} 
    
       \textbf{D-CLOT}$_{B}$ & w/o GC  & 57.2  & 40.6  & 18.6 & 51.1 & 34.9 & 23.8 & 49.2   & 43.7&30.5 &48.5  &58.0 & 35.9 & 59.0&63.6 &40.0\\ 
           & w/o GC in $F_{R}$  &  42.0 & 32.1  & 18.4 &53.0 & 36.1  & 25.3 &  41.6 & 38.7& 25.2&  47.8& 57.6& 35.0 &65.2 &68.9 &44.8\\
            &w/o GC in $F$ & 56.3  & 39.5  & 18.1 & 53.8 & 36.2  & 25.7 & 47.0  &44.0 &28.5 & 54.9 &66.6 & 39.0 & 68.1 &68.7 &45.7\\
         \rowcolor{NavyBlue!13}
\cellcolor{white} & Full &58.5 & 41.3  & 19.1  & 54.9   & 36.8 & 26.4  & 50.8 & 47.0 & 32.0 & 64.9 & 72.1 & 40.4 &  70.4 & 78.2 & 51.8 \\

        \bottomrule     
    
        \end{tabular}}

    \label{tab:ablation_study},

\end{table*}
\begin{figure*}[t]
    \centering   
    \includegraphics[width=0.85\linewidth]{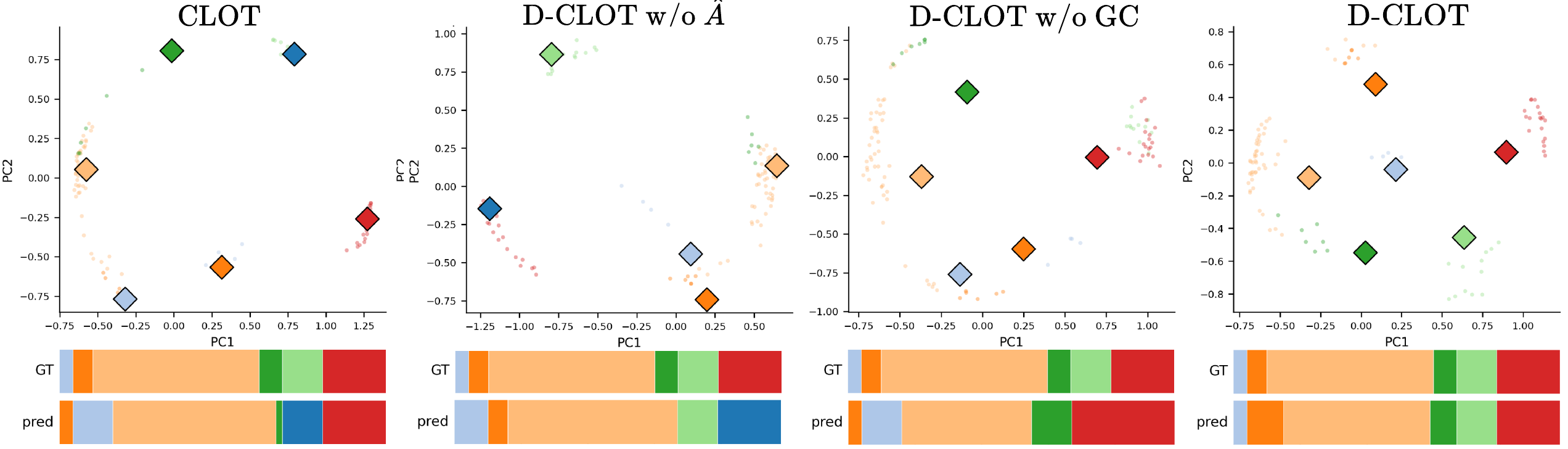}    
     
    \caption{\footnotesize\textbf{Qualitative ablation of the D-CLOT components on the prototype geometry and the resulting segmentation.} For a representative YTI video we show, top, the first two principal components of the refined frame embeddings $\mathbf{F}_R$ (dots, coloured by ground-truth action) together with the action prototypes $\mathbf{A}$ (diamonds) and, bottom, the ground-truth (GT) and predicted (pred) frame labels. The four panels ablate the two modules: (i) the CLOT baseline; (ii) D-CLOT without prototype re-estimation (w/o $\hat{A}$); (iii) D-CLOT without the graph constraint (w/o GC); and (iv) the full D-CLOT model. In CLOT, two prototypes collapse onto the same region and a spurious action (\textcolor{blue}{blue}) is predicted where none exists in the ground truth. The graph constraint restores temporally coherent boundaries but leaves the spurious prototype active; prototype re-estimation relocates the prototypes towards their supporting frame clusters. Only the full model, combining both, removes the spurious assignment and recovers a segmentation that closely matches the ground truth. Colours denote action classes.}
\label{fig:qualitative_ablation}
    \vspace{-1.5em}
\end{figure*}

\subsection{Assembly101: a fine-grained procediral benchmark}
\label{sec:assembly}
Table~\ref{tab:dataset-stats} highlights the difficulty of our Assembly101 subset. It contains the longest videos on average ($13{,}048$ frames), the largest activity diversity ($15$ toy categories with $11$--$42$ actions each), and substantial background ($42.1\%$). Together with its multi-view recordings and V-JEPA2 features, these properties make it more challenging than existing cooking and instructional benchmarks.

Table~\ref{tab:assembly101_baseline} reports the first unsupervised action-segmentation results on Assembly101. Both D-CLOT and D-CLOT$_{B}$ improve substantially over CLOT in F1 and mIoU at the video level, at the cost of a small MoF drop, again reflecting the MoF/F1--mIoU trade-off under the long-tailed, fine-grained action distribution of this benchmark ($11$--$42$ actions per toy category). Activity-level performance is markedly lower for every method, confirming that Assembly101 is a harder benchmark than the cooking and instructional datasets, owing to its larger action vocabulary and multi-view structure. D-CLOT$_{B}$ obtains the best overall average, indicating that graph-constrained action refinement generalises beyond the cooking and instructional domains to large-scale, fine-grained assembly video. Our method provides the strongest first baseline, but the scores indicate substantial room for future work.

\begin{figure}[t]
\centering
\vspace{-3em}
\begin{minipage}[t]{0.47\columnwidth}
    \centering
    \hspace*{-0.3\linewidth}
    \includegraphics[width=1.65\linewidth]{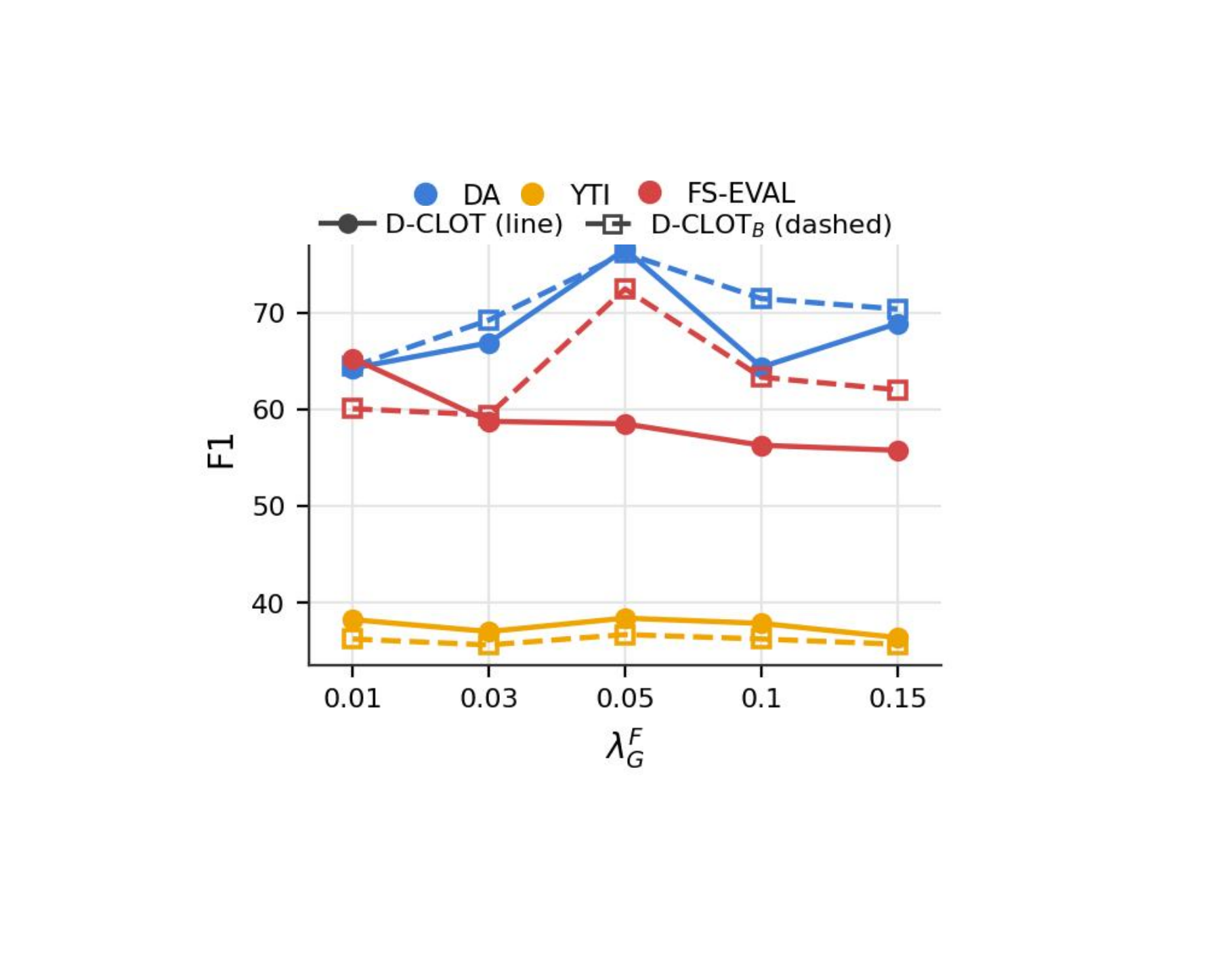}
    \vspace{-15mm}
    
    {\footnotesize (a) Frame graph weight $\lambda_G^F$}
    \vspace{-2em}
\end{minipage}
\begin{minipage}[t]{0.47\columnwidth}
    \centering
    \hspace*{-0.25\linewidth}
    \includegraphics[width=1.65\linewidth]{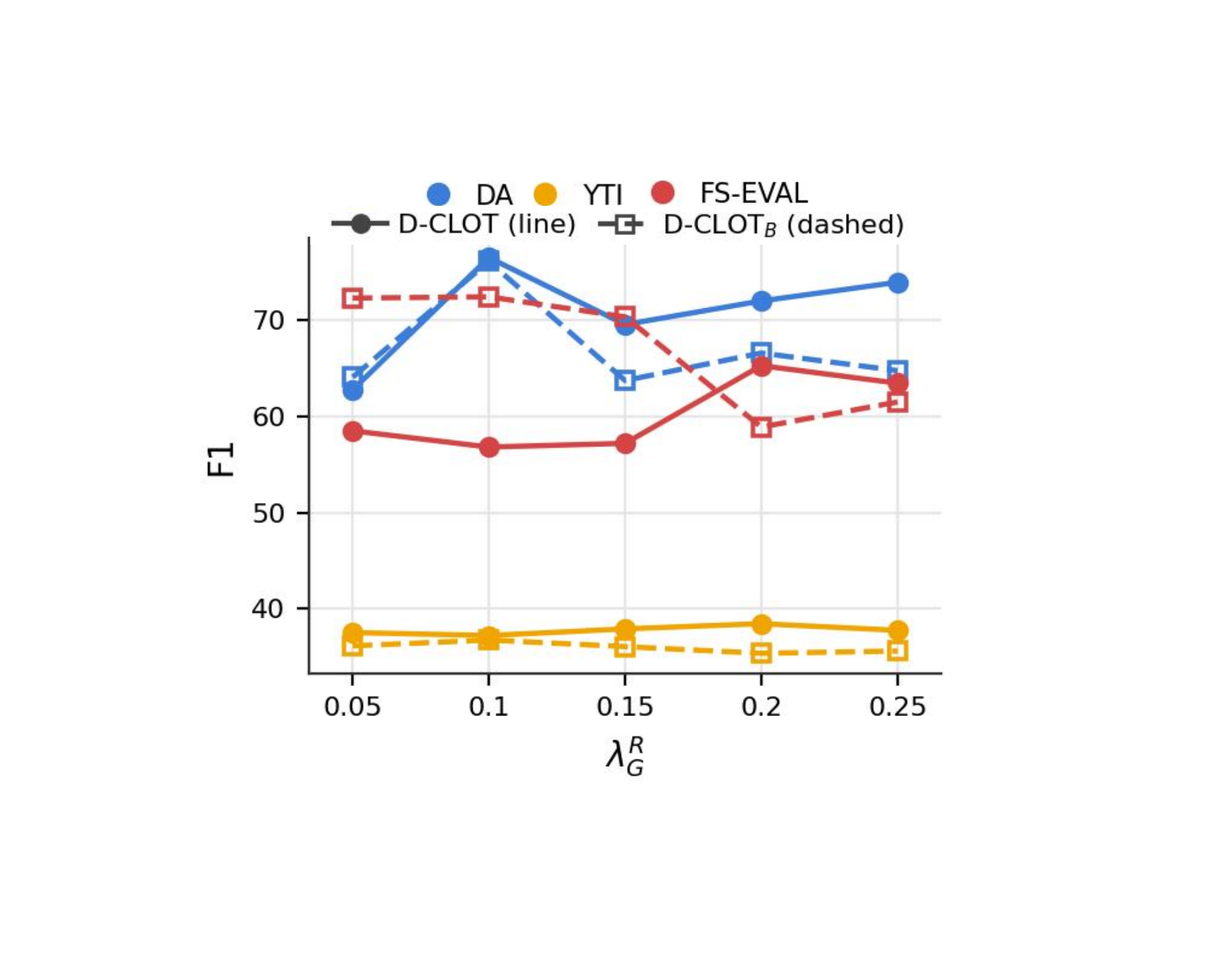}
    \vspace{-15mm}
    
    {\footnotesize (b) Refine graph weight $\lambda_G^R$}
    \vspace{-2em}
\end{minipage}
\vspace{-3em}

\begin{minipage}[t]{0.47\columnwidth}
    \centering
    \hspace*{-0.25\linewidth}
    \includegraphics[width=1.65\linewidth]{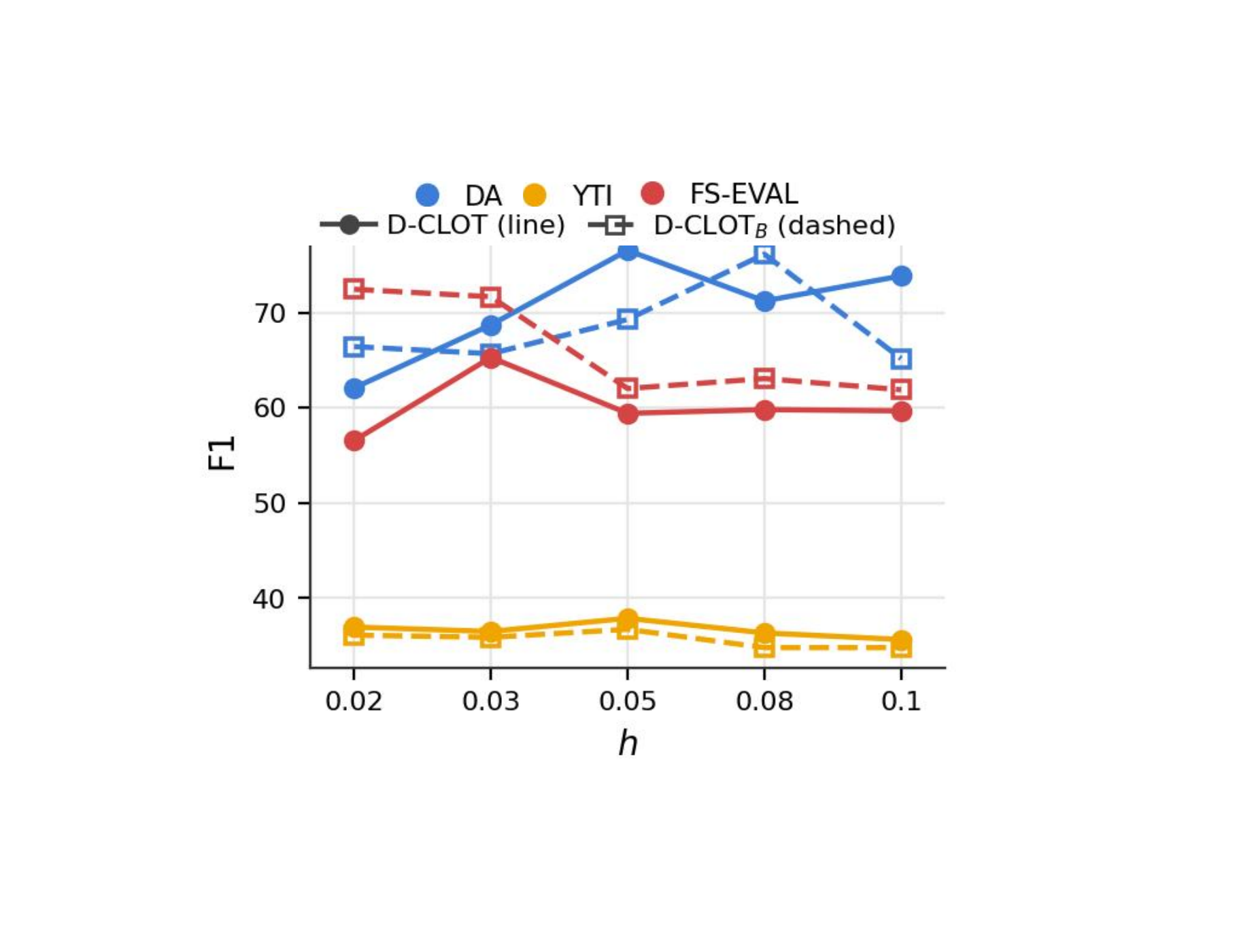}
    \vspace{-15mm}
    
    {\footnotesize (c) Graph bandwidth $h$}
    \vspace{-3.5em}
\end{minipage}
\hspace{-0.025\columnwidth}
\begin{minipage}[t]{0.47\columnwidth}
    \centering
    \hspace*{-0.25\linewidth}
    \includegraphics[width=1.65\linewidth]{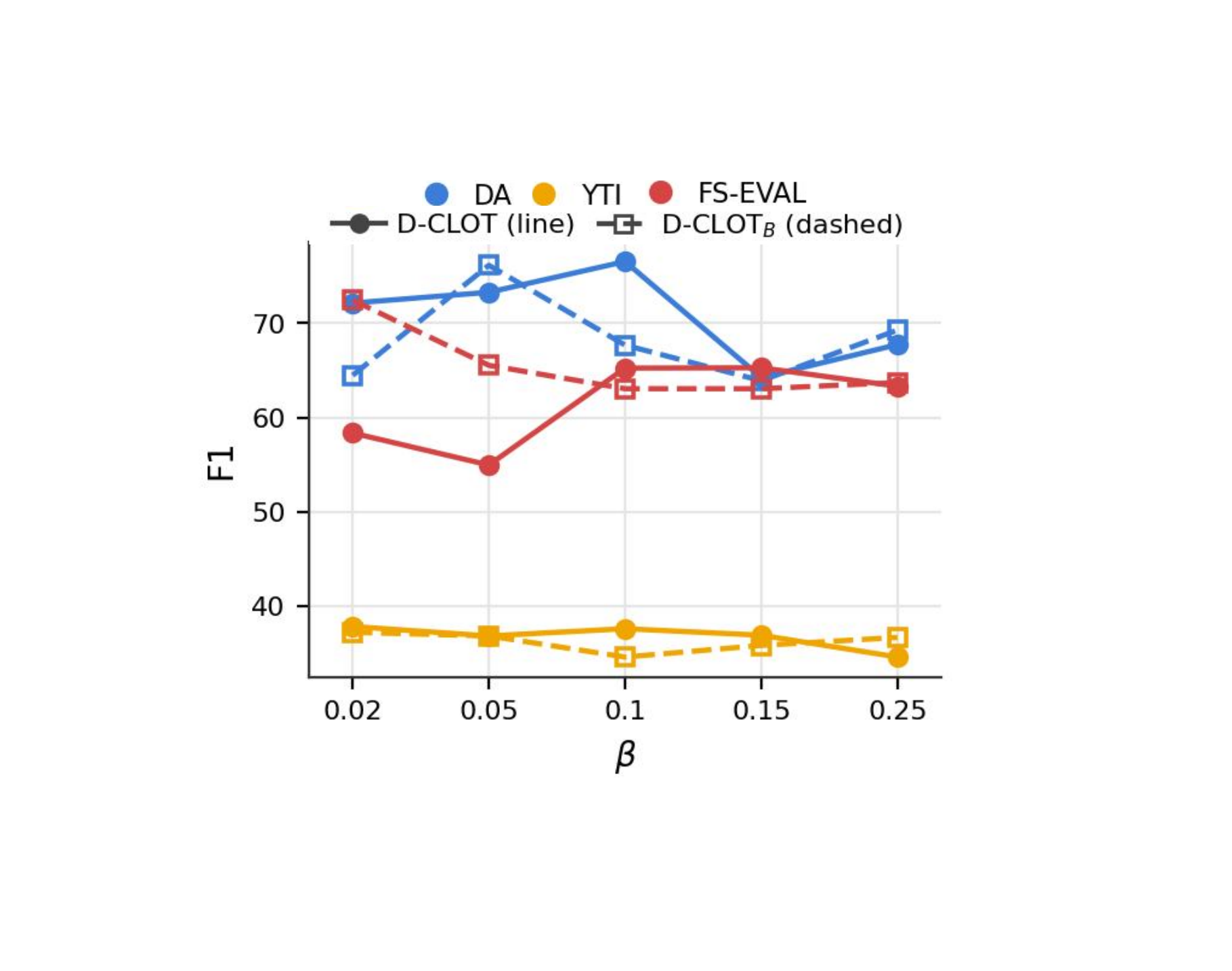}
    \vspace{-15mm}
    
    {\footnotesize (d) Blend factor $\beta$}
    \vspace{-3.5em}
\end{minipage}
\vspace{-3em}

\begin{minipage}[t]{0.47\columnwidth}
    \centering
    \hspace*{-0.25\linewidth}
    \includegraphics[width=1.5\linewidth]{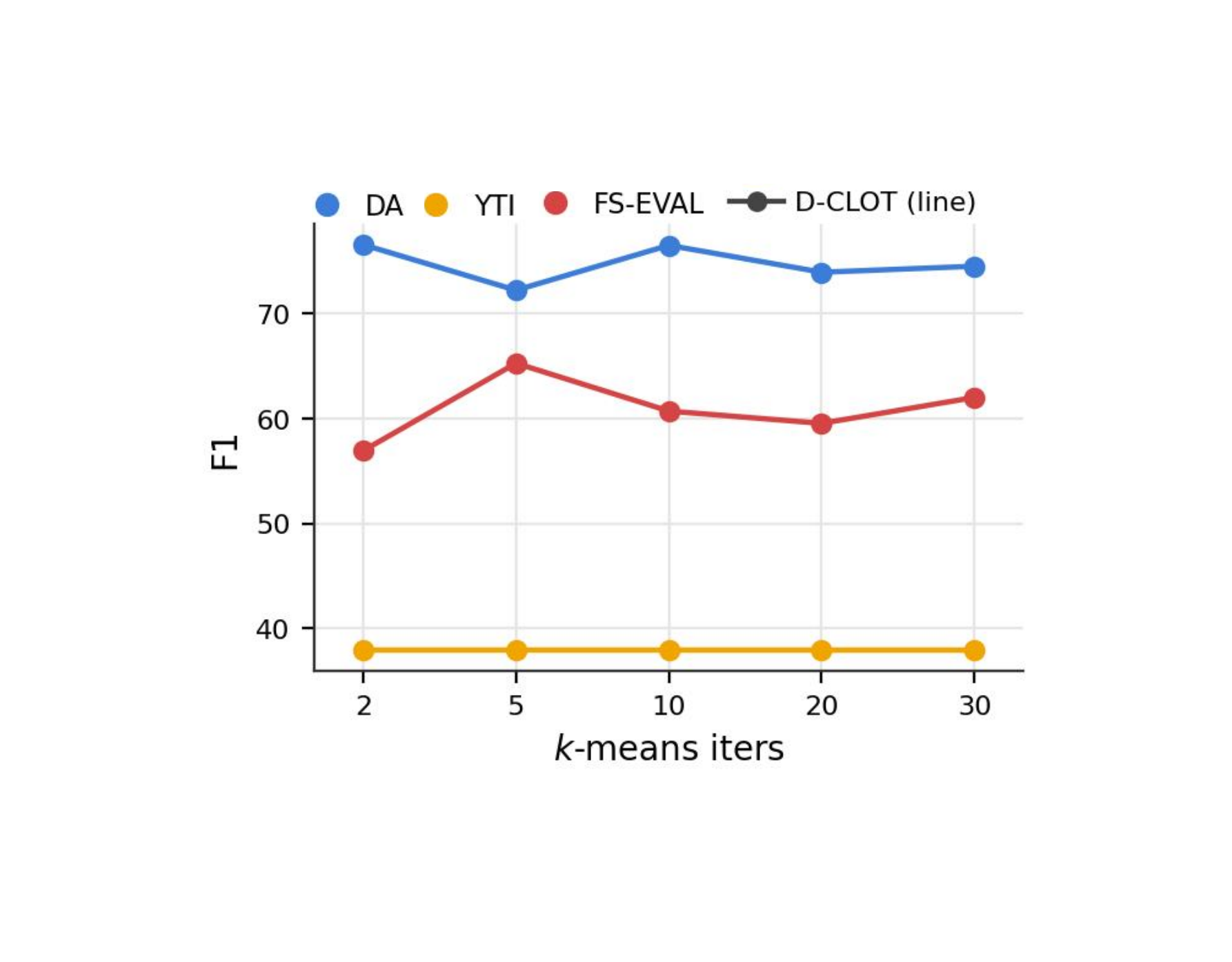}
    \vspace{-15mm}
    
    {\footnotesize (e) $k$-means refinement}
    \vspace{-3em}
\end{minipage}
\hspace{-0.025\columnwidth}
\begin{minipage}[t]{0.47\columnwidth}
    \centering
    \hspace*{-0.25\linewidth}
    \includegraphics[width=1.5\linewidth]{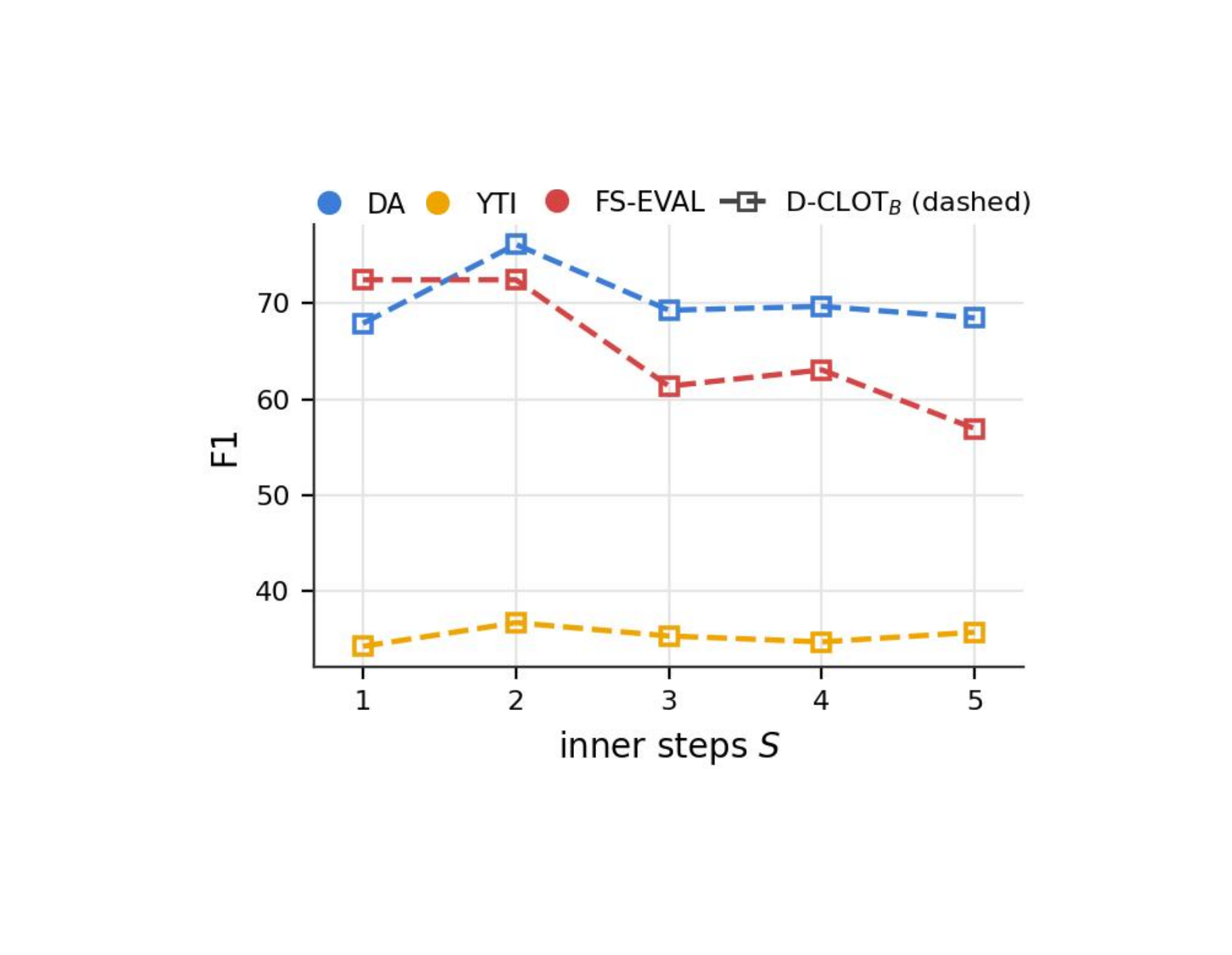}
    \vspace{-15mm}
    
    {\footnotesize (f) Barycentric refinement}
    \vspace{-3em}
\end{minipage}

\vspace{-2em}
\caption{\footnotesize Sensitivity analysis on F1. Each panel varies one hyperparameter while keeping the remaining configuration fixed. 
Colours denote datasets and line styles denote D-CLOT and D-CLOT$_{B}$.
}
\vspace{-1.5em}

\label{fig:sensitivity_f1}
\end{figure}
\subsection{Ablation study}
\label{sec:ablation}
Table~\ref{tab:ablation_study} isolates the contribution of the three proposed components: graph regularization on the frame embeddings ($\tilde F$), graph regularization on the refined frame embeddings ($\tilde F_R$), and action-embedding refinement ($A$). These components play complementary roles. The graph on $\tilde F$ preserves local visual neighborhoods before segment-level propagation, whereas the graph on $\tilde F_R$ regularizes a representation already enriched by cross-attention and therefore acts more directly on temporal coherence.

The results show that action refinement is most effective when applied to a graph-stabilized representation. Updating $A$ alone is unstable for D-CLOT, since prototypes are re-anchored to an unconstrained refined space. Conversely, graph regularization alone yields dataset-dependent gains, but does not consistently match the full configuration. Overall, the full model improves over CLOT on most dataset/metric pairs, especially in F1 and mIoU, showing that geometry stabilization and prototype refinement address different aspects of the representation--prototype mismatch.

A few ablated configurations outperform the full D-CLOT model on metrics. These cases reflect metric-specific trade-offs rather than a failure of the full design. On FS-Mid, removing $A$ slightly increases MoF, but reduces both F1 and mIoU, consistent with MoF's bias toward dominant long-duration actions. On FS-Eval, using $\tilde F_R$ and $A$ without $\tilde F$ improves F1, but causes a large MoF drop and slightly lower mIoU, indicating boundary-level gains at the expense of frame coverage. In contrast, D-CLOT$_{B}$ shows a more stable pattern: its full configuration dominates all partial variants across datasets and metrics. This supports the assignment-aware barycentric update as a more robust refinement mechanism.

Fig.~\ref{fig:qualitative_ablation} illustrates the complementary roles of graph regularization (GC) and action refinement. CLOT disrupts the action order, while GC recovers more coherent temporal boundaries. However, GC still assigns frames to a spurious action prototype, shown by the blue segment. Updating $A$ corrects this prototype mismatch: in the full D-CLOT model, the blue segment disappears, indicating that any frame no longer selects the corresponding prototype. Thus, GC mainly regularizes temporal structure, whereas action refinement corrects the prototype-to-frame assignment. On this video, D-CLOT improves MoF from $0.663$ to $0.939$, F1 from $0.333$ to $1.000$, and mIoU from $0.348$ to $0.883$.

The PCA plot supports this interpretation. Stable prototypes, such as the light-green one, remain almost unchanged, while the dark-green prototype moves substantially toward the relevant frame cluster. This displacement reduces the influence of the dark-blue prototype and resolves the erroneous assignment. The proximity between the light-blue and dark-orange prototypes also explains the observed confusion in ordering and boundaries in CLOT. Overall, the figure shows that D-CLOT improves segmentation by jointly correcting temporal structure and prototype geometry.

\subsection{Sensitivity analysis}
\label{subsec:sensitivity_analysis}
We analyse the sensitivity of D-CLOT and D-CLOT$_{B}$ with respect to the main hyperparameters introduced by the graph-constrained stabilisation and action-refinement modules in Fig.~\ref{fig:sensitivity_f1}. Since temporal action segmentation is highly sensitive to over-segmentation and boundary fragmentation, we report F1 in the main paper and provide the corresponding MOF and mIoU plots in the supplementary material. Compared with MOF, which can be biased towards frequent or long-duration actions, and compared with mIoU, it is less brittle under the noisy pseudo-labels inherent to unsupervised training. F1 provides a more diagnostic measure of segmental quality.  
\noindent\textbf{Graph fidelity weights ($\lambda_G^F$, $\lambda_G^R$).}
On DA, F1 peaks sharply at $\lambda_G^F=0.05$ and $\lambda_G^R=0.1$ for both variants: a moderate amount of graph regularisation stabilises the refined frame geometry, while excessive regularisation over-constrains it, as expected from~\eqref{eq:graph_loss}. D-CLOT$_{B}$ is smoother and less sharply peaked than D-CLOT, particularly with respect to $\lambda_G^F$. On YTI, F1 varies little: with 62\% background frames, there is limited geometry to anchor to, so the modules act as a stabiliser rather than a strong amplifier, the intended fallback behaviour, not a shortcoming. FS-Eval sits in between, but the barycentric variant benefits from the refined graph weight, though this gain requires controlled strength to avoid over-smoothing.

\noindent\textbf{Graph bandwidth ($h$).}
On DA, small $h$ collapses the affinity graph to near-diagonal and the term stops regularising, while a large $h$ flattens it and discards local structure, per~\eqref{eq:affinity}. YTI stays flat, consistent with a stabilisation role rather than a performance lever under weak temporal regularity. FS-Eval is mixed: flat for D-CLOT, but a 10-point swing for D-CLOT$_{B}$, again reflecting the barycentric variant's greater ability to exploit graph structure when it is informative.

\noindent\textbf{Blend factor ($\beta$).}
On DA, moderate blending helps both variants: the transport plan is reliable, so re-anchoring prototypes captures real structure. On YTI, a small blending is best, with only mild degradation as $\beta$ grows; this is not a failure of the update but confirmation that the bounded, convex blend of~\eqref{eq:blend} degrades gracefully under unreliable transport mass. FS-Eval is less consistent between variants: D-CLOT peaks at the largest $\beta$ tested, D-CLOT$_{B}$ barely moves.\vspace{1em}

\vspace{-0.5em}
\noindent\textbf{Stabilisation budget ($k$-means iterations, inner steps $S$).}
D-CLOT is robust to its $k$-means budget~\eqref{eq:kmeans_update}: F1 does not change on YTI, and our default is within 0.1 point of the best value on DA. Therefore, sharper prototype relocation is beneficial here without requiring careful tuning. 
D-CLOT$_{B}$'s inner-step count $S$~\eqref{eq:aux_inner} matters more. The three datasets peak sharply at $S=2$, our default. In general, more steps pull the representation harder toward the encoder anchor, which helps denoise YTI's weaker signal but risks over-smoothing FS-Eval's already coarse, 12-class boundaries. We fix $S=2$, which matches DA and YTI exactly and balances the other dataset complementary preferences.

Across datasets, the sensitivity curves demonstrate that graph-constrained action refinement is robust to moderate hyperparameter variations. No parameter causes a substantial performance degradation, and the best results generally arise from intermediate graph regularization and conservative prototype updates. This suggests that refinement improves the geometry of frame representations without excessively constraining OT-based action discovery. D-CLOT is preferable when stronger $k$-means prototype relocation is beneficial, whereas D-CLOT$_{B}$ yields smoother, assignment-aware updates through its auxiliary graph-stabilized representation. The lower sensitivity observed on YTI indicates that, for noisier videos pooled across activities, refinement primarily stabilizes learning rather than substantially amplifying performance. This behavior is particularly evident for $\beta$: the bounded update in ~\eqref{eq:blend} limits prototype changes on YTI while exploiting clearer structure on DA. Overall, these results support the selected default configuration and show that the improvements do not depend on narrowly tuned hyperparameters.

\begin{figure}[t]
    \centering
   \begin{minipage}[t]{ \columnwidth}
    \centering
    \vspace{-1.2em}
    \includegraphics[width=0.95\linewidth,trim={0 50 0 50},clip]{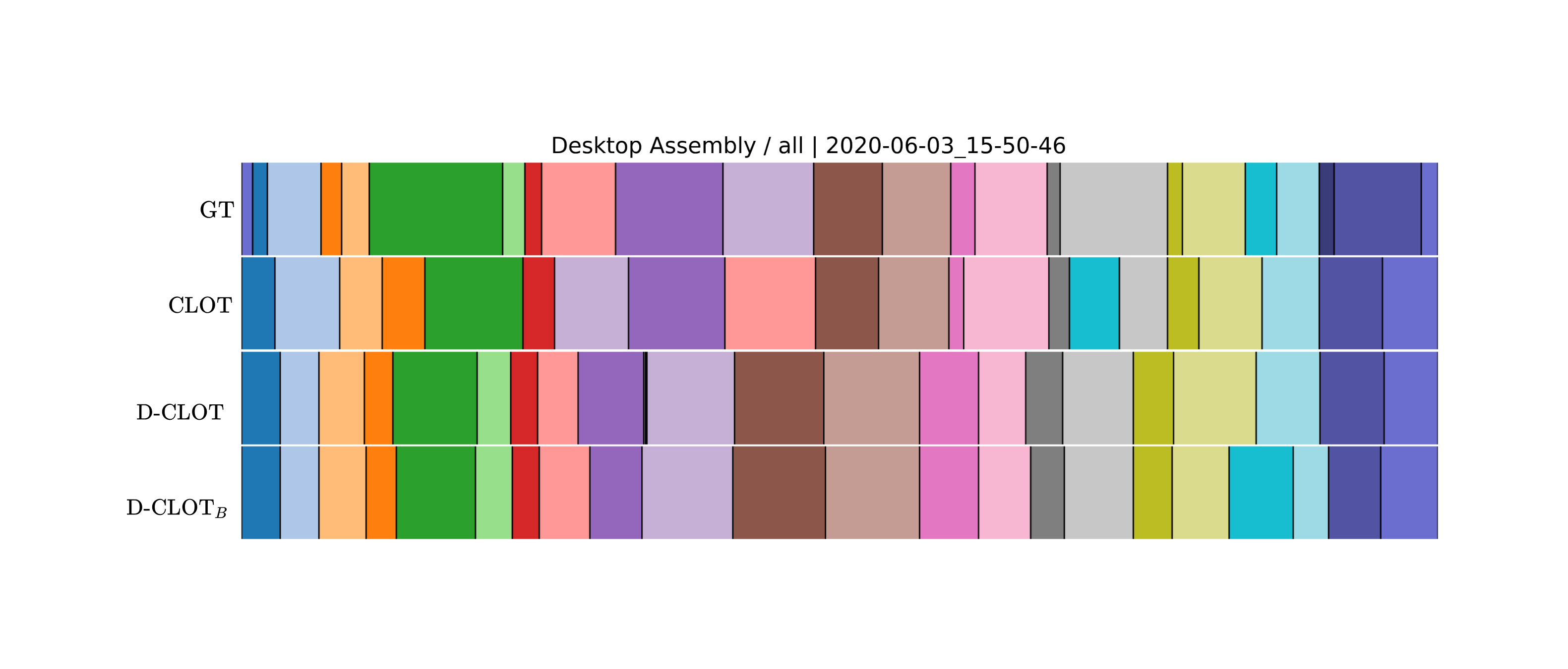}

\end{minipage} 
\begin{minipage}[t]{ \columnwidth}
    \centering
    \vspace{-1.5em}
    \includegraphics[width=0.95\linewidth,trim={0 50 0 50},clip]{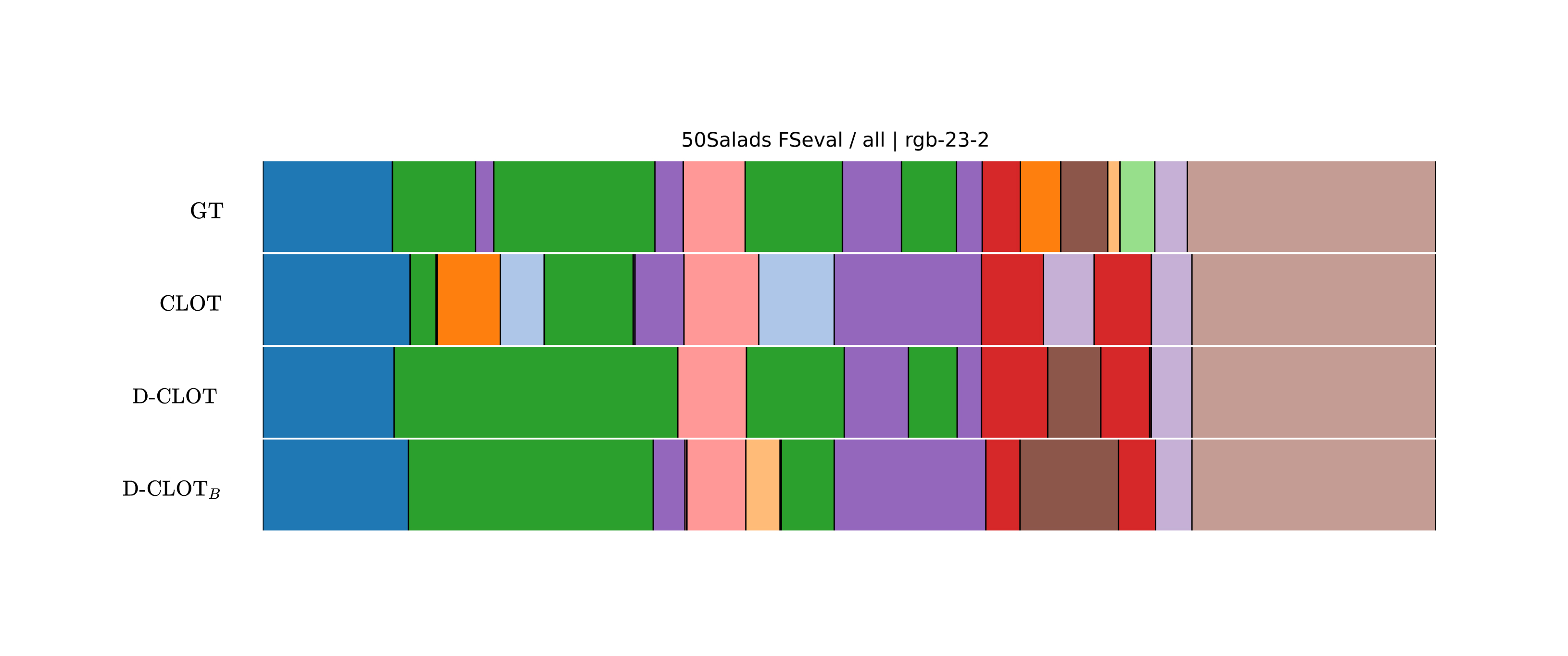}
    \vspace{-1em}   
\end{minipage} 
    \par 

\caption{\footnotesize Example segmentation from DA\cite{Kumar22} (top) and FS (Eval)\cite{50salads} (bottom). We report the ground-truth (GT), the results of CLOT, and the result in ours two variants, D-CLOT and D-CLOT$_{B}$.}

\vspace*{-4mm}
 \label{fig:qualitative_segmentation}
\end{figure}
\subsection{Qualitative results.}
Figure~\ref{fig:qualitative_segmentation} compares the temporal segmentations produced by CLOT, D-CLOT, and D-CLOT$_B$ on representative videos from 50Salads and Desktop Assembly. On FS-Eval, CLOT introduces spurious classes in the early--middle region and over-merges the central structure into a single block, losing the fine transitions. D-CLOT removes them and restores cleaner, more coherent segments, while D-CLOT$_B$ recovers short, minority actions and thin boundaries that the other rows miss. On DA, CLOT preserves the global action order but drifts boundaries and drops several short steps. D-CLOT tightens these boundaries, and D-CLOT$_B$ yields the closest alignment to GT, respecting the procedural order and recovering the short intermediate segments. Re-anchoring the prototypes to the graph-stabilized geometry thus fixes the spurious-class and boundary-merging failures of CLOT, and the assignment-aware update preserves exactly the short, imbalanced actions that drive F1/mIoU while barely affecting MoF.

\section{Discussion and Conclusion} 
\label{sec:conclusions}
We introduced D-CLOT, a graph-constrained extension of CLOT for unsupervised temporal action segmentation. The proposed framework addresses the mismatch between progressively refined frame representations and latent action prototypes by jointly regularizing the representation geometry and periodically updating the prototypes. We investigated two refinement strategies: a $k$-means update and an assignment-aware optimal transport barycentric update, while preserving the original CLOT formulation.

Extensive experiments on five benchmarks demonstrate consistent improvements over CLOT, with substantial gains in segment-level quality, particularly in F1 and mIoU. The ablation analysis shows that graph-based regularization and prototype refinement are complementary: graph regularization promotes temporally coherent representations, whereas prototype refinement maintains alignment between latent actions and the evolving feature space. Among the proposed variants, the assignment-aware barycentric refinement consistently provides the most robust performance across datasets.

Beyond improving the state of the art, we present the first unsupervised temporal action segmentation benchmark on Assembly101, highlighting the challenges posed by fine-grained actions, viewpoint variability, and procedural diversity. These results suggest that continuously adapting latent prototypes to an evolving representation space is a key ingredient for scalable unsupervised temporal action segmentation, and motivate future work on richer prototype learning and more expressive latent action models.

\vspace{-1em}

\section*{Acknowledgments}
This work was supported by grant \small{PRE2020-094714}, the project \small{PID2019-110977GA-I00} and the project \small{PID2023-151351NB-I00}, funded by Ministerio de Ciencia e Innovación (MCIN)/ Agencia Estatal de Investigación (AEI) /10.13039/501100011033, by European Social Fund (ESF) Investing in your future and by \textit{ERDF, UE}. It was also supported by an \textit{Alexander von Humboldt (AvH)} fellowship for experienced researchers funded by the \textit{AvH Foundation}.
\bibliographystyle{IEEEtran}
\bibliography{main}



\appendix
 
This supplementary material provides additional evidence supporting the robustness and practicality of D-CLOT and D-CLOT$_{\mathrm{B}}$. It examines how consistently the methods improve individual videos, rather than relying only on dataset-level averages (App.~\ref{sec:per_video_gains}). It also reports complete hyperparameter settings  (App.~\ref{sec:hyperparameter}), additional sensitivity results  (App.~\ref{sec:sensitive_analysis_mof_miou}), computational-cost measurements  (App.~\ref{sec:computational}), and further qualitative examples  (App.~\ref{sec:more_examples}). Together, these analyses clarify where the gains come from, how stable they are, and the additional cost of each variant.

\appendices
\section{Where Do the Gains Come From?}
\label{sec:per_video_gains}

\begin{figure}[t]
    \centering
    
    \includegraphics[width=0.85\linewidth]{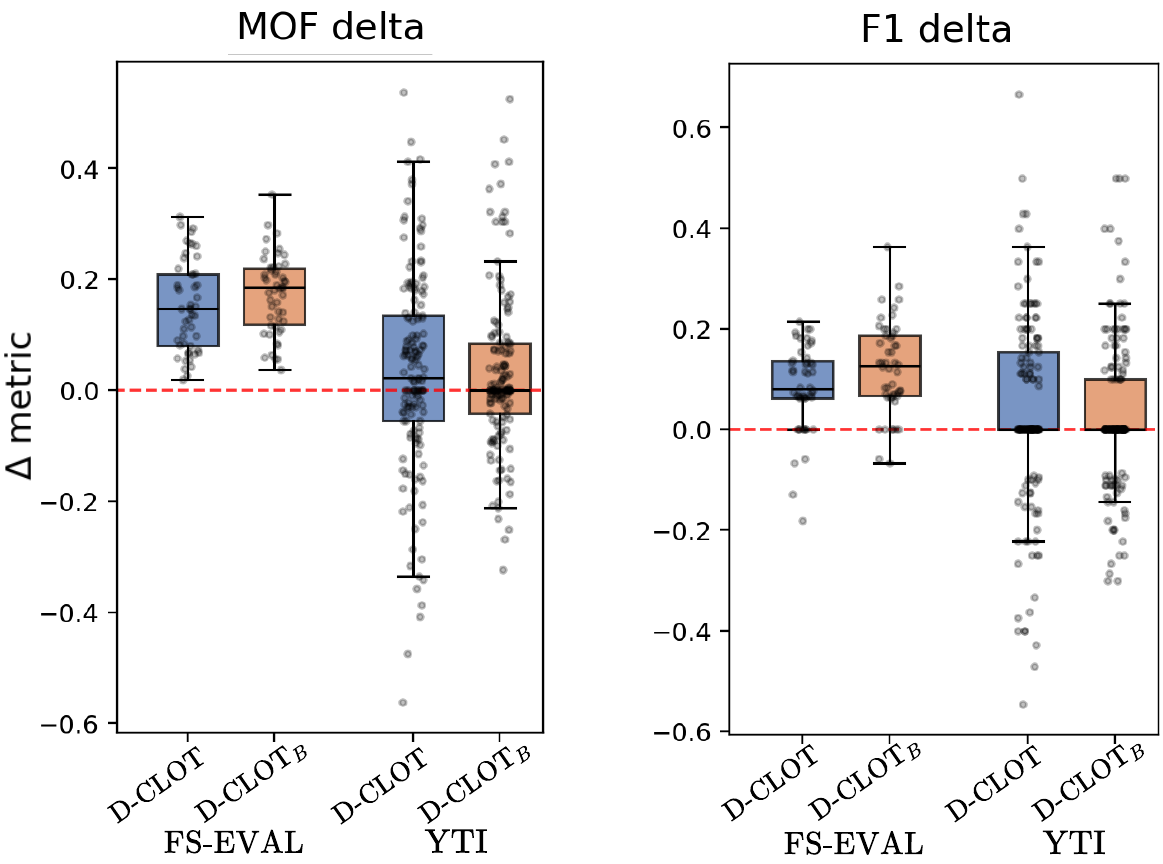}
   \caption{ \footnotesize
    Per-video improvement over CLOT~\cite{bueno-benito2025} for MoF and F1 on FS-Eval ($n{=}50$) and YTI ($n{=}149$, pooled over its five activities). For each video, $\Delta=\text{variant}-\text{CLOT}$ is computed separately for D-CLOT and D-CLOT$_{B}$; each box shows the median, interquartile range, and whiskers of $\Delta$, with per-video values overlaid as jittered points. Mass above the \textcolor{red}{dashed} $\Delta{=}0$ line marks videos that improve over CLOT, mass below it marks regressions. 
    \label{fig:yti_fseval_mof_f1_boxplot}
    } 
\end{figure}

Figure~\ref{fig:yti_fseval_mof_f1_boxplot} analyses \emph{where} the aggregate gains come from, moving from dataset means to the per-video distribution. A mass above it denotes videos that improve, and a mass below it denotes regressions. This view distinguishes a broad, population-level effect from one carried by a few sequences, which dataset averages alone cannot reveal.

On FSeval, the entire distribution shifts upward: for both variants, the boxes lie predominantly above the zero line in MoF and in F1 for D-CLOT/D-CLOT$_{B}$. The gains are therefore not driven by a few favourable sequences but reflect a consistent, population-level effect, and D-CLOT$_{B}$ shows the stronger shift, indicating that the assignment-aware barycentric update yields a more systematic alignment between frame representations and action prototypes.

YTI exhibits a complementary regime. Its distributions stay concentrated around zero but develop pronounced positive tails, producing sizeable average gains despite lower per-video win rates: the refinements do not improve every sequence, but where they help, the gains outweigh the occasional regressions. The contrast between the two datasets thus reveals two operating modes, a stable and broadly distributed improvement on FSeval and a selective but high-impact correction on YTI, the latter consistent with YTI's shorter, more heterogeneous videos and its independently trained activity-specific checkpoints, which introduce greater per-video variability.

\section{Hyperparameter Details}
\label{sec:hyperparameter}
D-CLOT and D-CLOT$_{B}$ share the same backbone and optimisation protocol, inherited unchanged from the underlying OT-based encoder: we use the Adam optimiser with a weight decay of $10^{-4}$, and, following~\cite{Xu2024, bueno-benito2025}, set the number of action prototypes $K$ to the ground-truth number of actions per activity category/dataset. The temporal-consistency radius $r$ of the base OT solver is fixed at $0.02$ for all datasets, except for Breakfast, where the longer and more finely segmented videos benefit from a wider window ($r=0.04$). The frame/action affinity graph used by the graph-constrained modules is sparsified to its $32$ nearest neighbours throughout, except Assembly101 ($16$), whose substantially longer, multi-view videos favour a sparser graph. For D-CLOT$_{B}$,  the auxiliary alignment weight is fixed at $\lambda_{\text{a}}=1.0$ for all datasets except Breakfast ($\lambda_{\text{a}}=0.25$), where a lighter alignment term better tolerates the noisier pseudo-labels typical of longer, unstructured activities.  

Beyond this shared backbone, D-CLOT and D-CLOT$_{B}$ introduce a small number of graph-regularisation and prototype-update hyperparameters: the frame and refined graph weights $\lambda_G^F$ and $\lambda_G^R$, the graph bandwidth $h$, and the blend factor $\beta$. Rather than treat these as free parameters tuned independently per dataset, we select them from the robust, wide-margin operating region identified by the sensitivity analysis in Sec.IV-E and in App.~\ref{sec:sensitive_analysis_mof_miou} which show performance to be stable under moderate perturbation of each of these values; the per-dataset choices used to produce our reported hyperparameters are given in Tables~\ref{tab:hparams_clot2} and~\ref{tab:hparams_clot2b} for D-CLOT and D-CLOT$_{B}$, respectively.
\begin{table}[t]
\centering
\small
\caption{Hyperparameter settings used for D-CLOT across datasets. FS (Mid), FS (Eval) and A101 denote the 50 Salads Mid, Eval splits, and Assembly101, respectively.}
    \setlength\tabcolsep{3pt}
    \resizebox{3in}{!}{
\begin{tabular}{lcccccc}
\toprule
 & BF & YTI & FS (Mid) & FS (Eval) & DA & A101 \\
\midrule
$\lambda_G^F$            &  0.10 & 0.10 & 0.05 & 0.05 & 0.05 & 0.03 \\
$\lambda_G^R$             & 0.10  & 0.15 & 0.10 & 0.20 & 0.10 & 0.05 \\
Graph bandwidth $h$          & 0.05 & 0.05 & 0.05 & 0.03 & 0.05 & 0.08 \\
Blend factor $\beta$         & 0.25  & 0.02 & 0.05 & 0.15 & 0.10 & 0.25 \\
K-means iterations           & 20  & 20   & 20   & 5    & 10   & 20 \\
Epochs                       & 25  & 15   & 60   & 20   & 110  & 35 \\
\bottomrule
\end{tabular} }

\label{tab:hparams_clot2}
\end{table}

\begin{table}[t]
\centering

\caption{Hyperparameter settings used for D-CLOT$_{B}$ across datasets. FS (M), FS (E) and A101 denote the 50 Salads Mid, Eval splits, and Assembly101, respectively.}
\small

    \setlength\tabcolsep{3pt}
    \resizebox{3in}{!}{
\begin{tabular}{lcccccc}
\toprule
 & BF & YTI & FS (Mid) & FS (Eval) & DA & A101 \\
\midrule
$\lambda_G^F$                & 0.01 & 0.05 & 0.10 & 0.05 & 0.05 & 0.03 \\
$\lambda_G^R$                 & 0.15 & 0.10 & 0.10 & 0.10 & 0.10 & 0.05 \\
Graph bandwidth $h$              & 0.02 & 0.08 & 0.05 & 0.02 & 0.08 & 0.08 \\
Blend factor $\beta$             & 0.05 & 0.25 & 0.25 & 0.02 & 0.05 & 0.25 \\
Barycentric inner steps $S$      & 2    & 2    & 2    & 2    & 2    & 2 \\
Epochs                           & 50   & 30  & 35   & 20   & 90  & 35 \\
\bottomrule
\end{tabular}}
\label{tab:hparams_clot2b}
\end{table}

\section{Sensitive Analysis}
\label{sec:sensitive_analysis_mof_miou} 
This section complements Section IV-E with MoF and mIoU for the same datasets. MoF and mIoU peak at the same hyperparameter value as F1 in almost every case. A few FS-Eval configurations shift the MoF optimum by one grid step and under $1.5$ points, with F1 and mIoU unaffected. This is consistent with FS-Eval's coarse $12$-class structure, where MoF depends on coverage of a few long segments and is sensitive to the handful of boundary frames a one-step change reassigns, whereas F1 and mIoU are not. No hyperparameter trades one metric off against another, so the F1 gains in the main text are not an artefact of over- or under-segmentation
\begin{figure}[t]
\centering
\vspace{-3em}
\begin{minipage}[t]{0.47\columnwidth}
    \centering
    \hspace*{-0.25\linewidth}
    \includegraphics[width=1.65\linewidth]{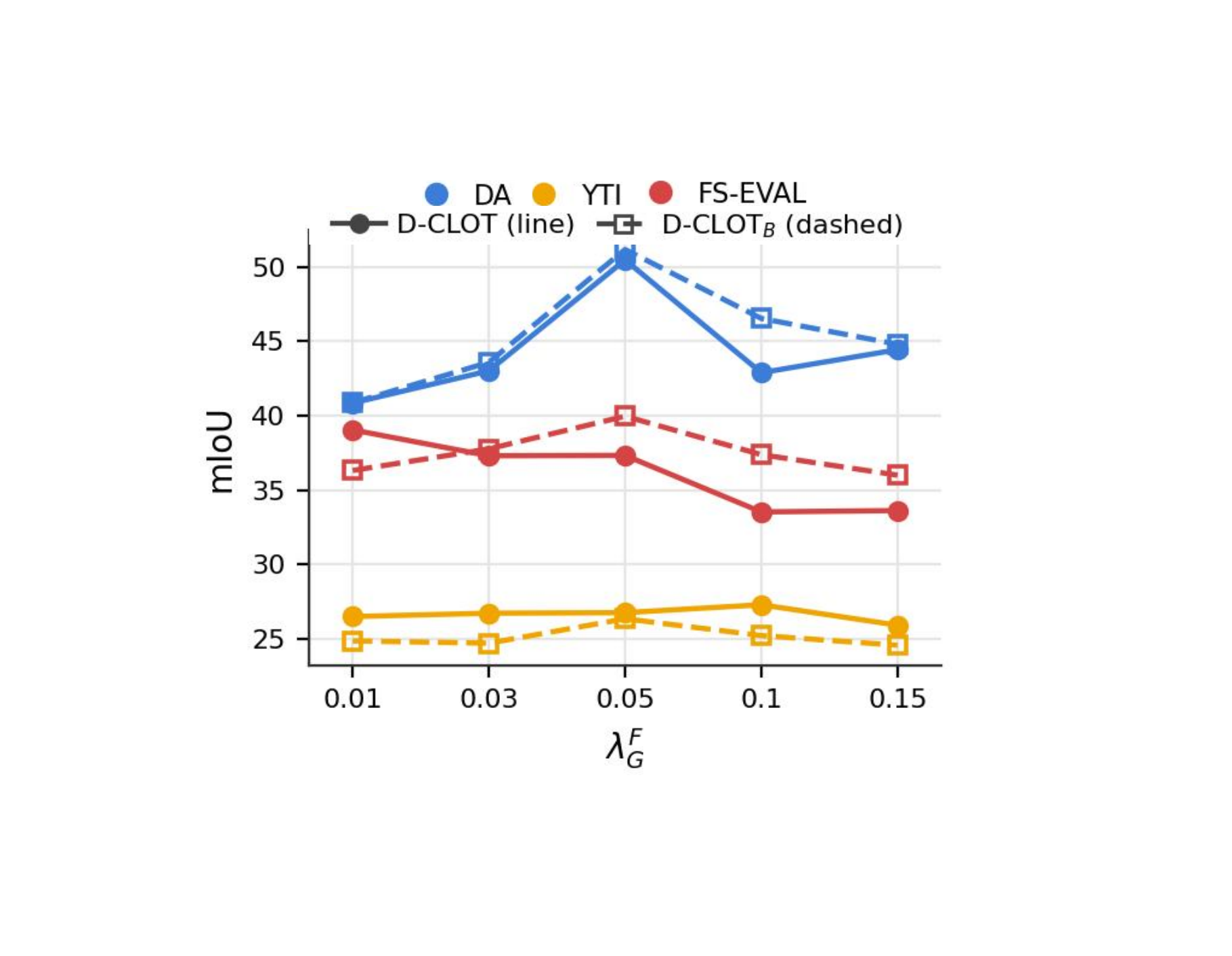}
    \vspace{-15mm}
    
    {\footnotesize (a) Frame graph weight $\lambda_G^F$}
    \vspace{-2em}
\end{minipage}
\begin{minipage}[t]{0.47\columnwidth}
    \centering
    \hspace*{-0.25\linewidth}
    \includegraphics[width=1.65\linewidth]{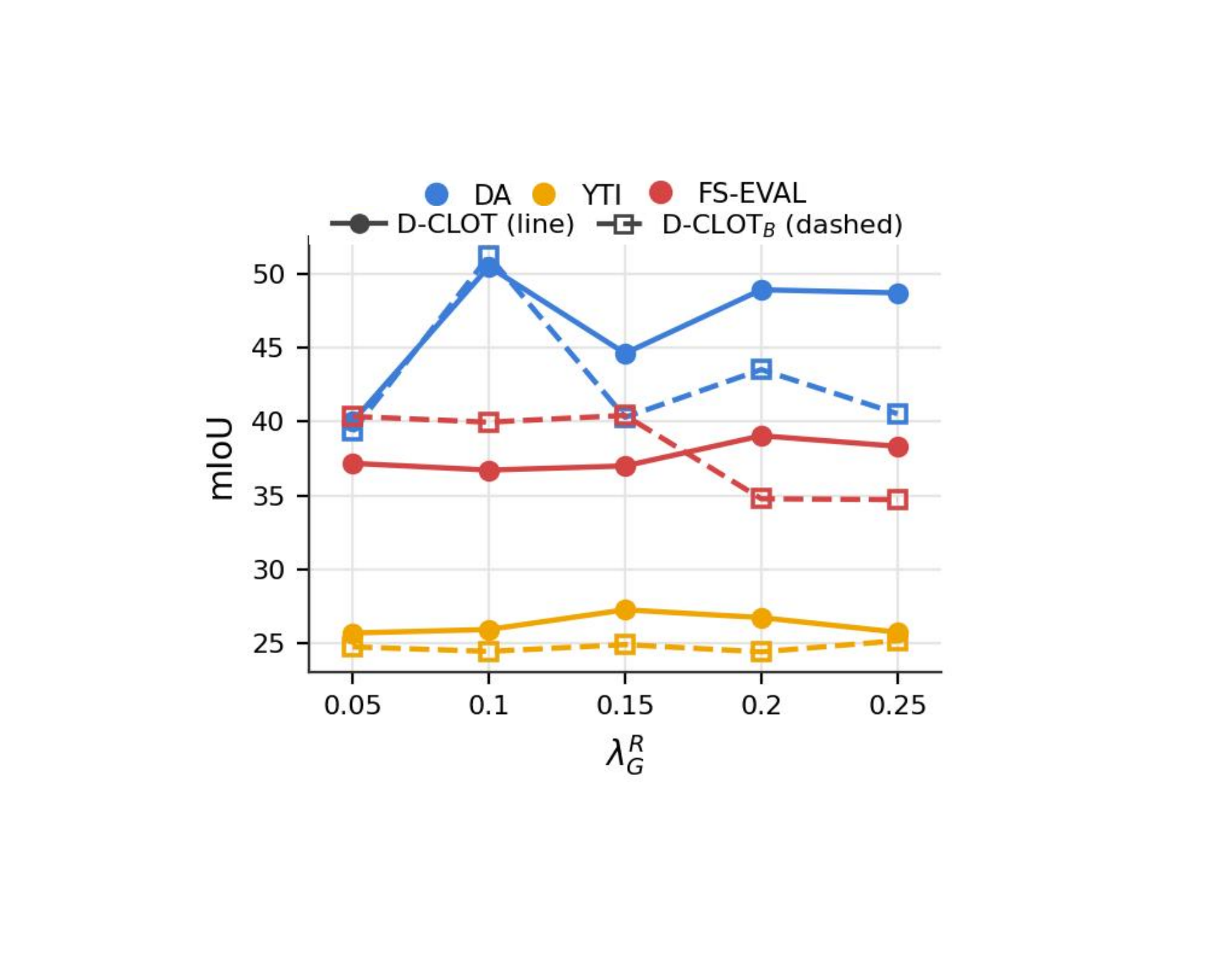}
    \vspace{-15mm}
    
    {\footnotesize (b) Refine graph weight $\lambda_G^R$}
    \vspace{-2em}
\end{minipage}
\vspace{-3em}

\begin{minipage}[t]{0.47\columnwidth}
    \centering
    \hspace*{-0.25\linewidth}
    \includegraphics[width=1.65\linewidth]{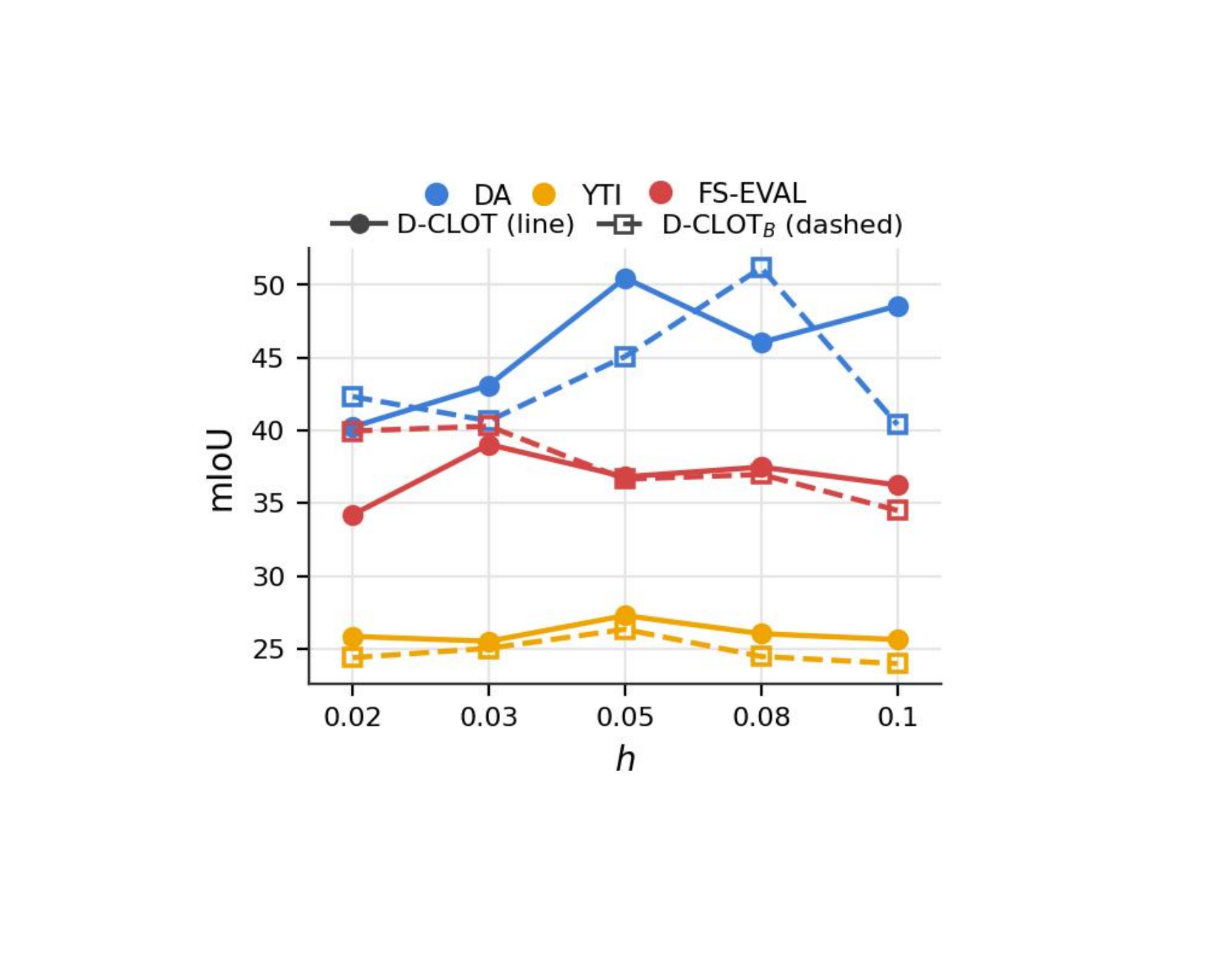}
    \vspace{-15mm}
    
    {\footnotesize (c) Graph bandwidth $h$}
    \vspace{-2em}
\end{minipage}
\hspace{-0.025\columnwidth}
\begin{minipage}[t]{0.47\columnwidth}
    \centering
    \hspace*{-0.25\linewidth}
    \includegraphics[width=1.65\linewidth]{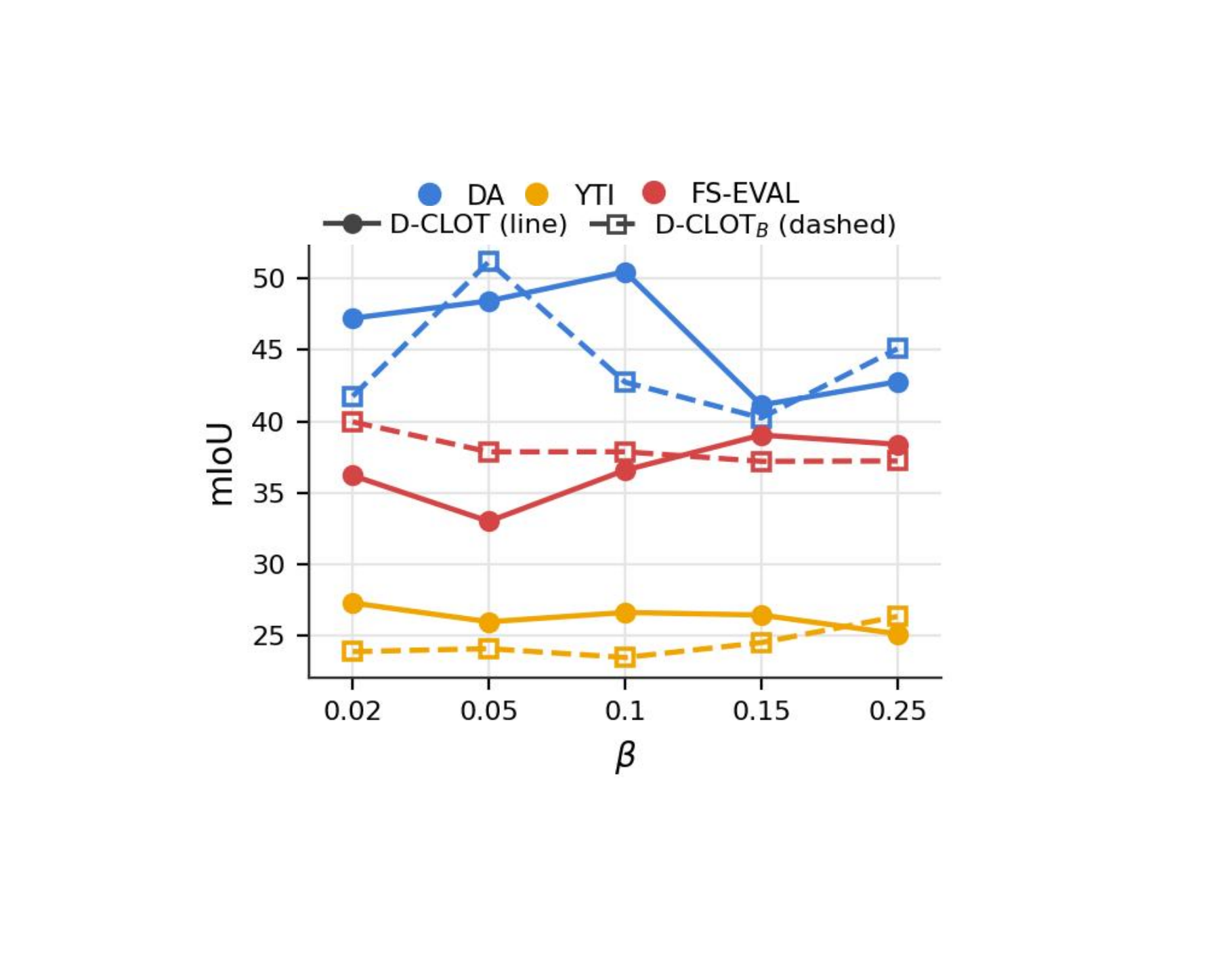}
    \vspace{-15mm}
    
    {\footnotesize (d) Blend factor $\beta$}
    \vspace{-2em}
\end{minipage}
\vspace{-3em}

\begin{minipage}[t]{0.47\columnwidth}
    \centering
    \hspace*{-0.25\linewidth}
    \includegraphics[width=1.5\linewidth]{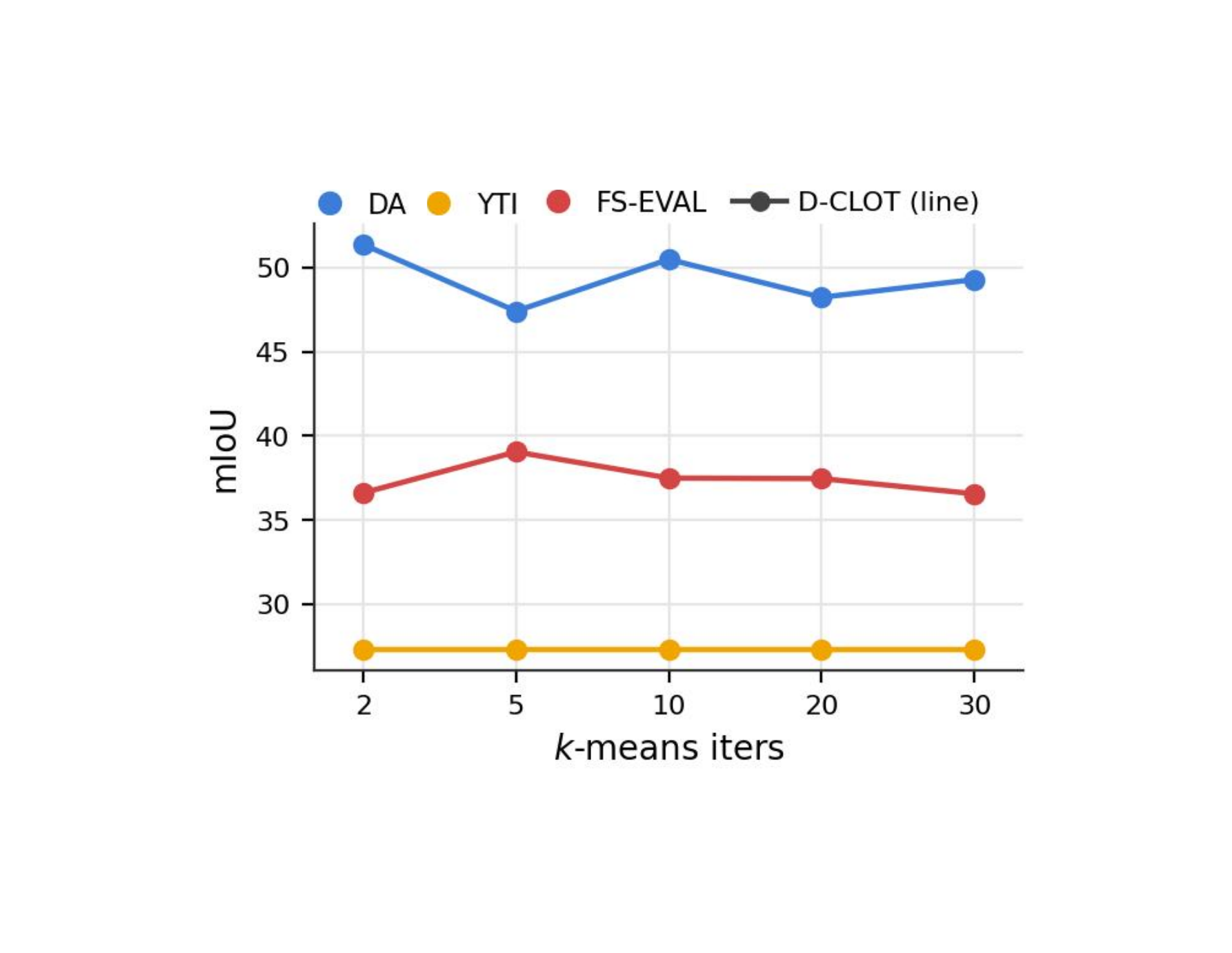}
    \vspace{-15mm}
    
    {\footnotesize (e) $k$-means refinement}
    \vspace{-2em}
\end{minipage}
\hspace{-0.025\columnwidth}
\begin{minipage}[t]{0.47\columnwidth}
    \centering
    \hspace*{-0.25\linewidth}
    \includegraphics[width=1.5\linewidth]{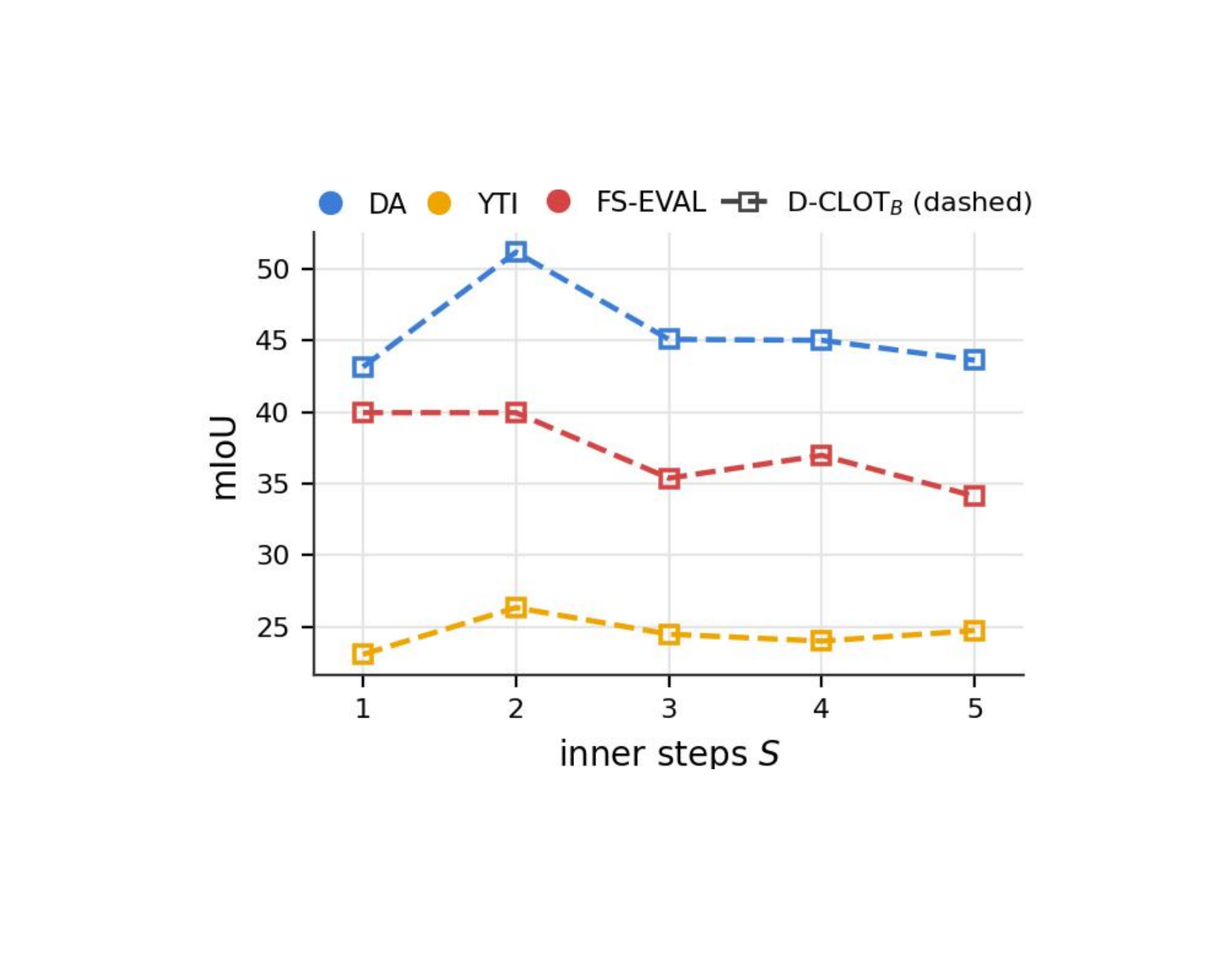}
    \vspace{-15mm}
    
    {\footnotesize (f) Barycentric refinement}
    \vspace{-2em}
\end{minipage}

\vspace{-1mm}
\caption{
Sensitivity analysis on mIoU. Each panel varies one hyperparameter while keeping the remaining configuration fixed. 
Colours denote datasets and line styles denote D-CLOT and D-CLOT$_{B}$.
}
\label{fig:sensitivity_miou}
\end{figure}
\begin{figure}[t]
\centering
\vspace{-3em}
\begin{minipage}[t]{0.47\columnwidth}
    \centering
    \hspace*{-0.25\linewidth}
    \includegraphics[width=1.65\linewidth]{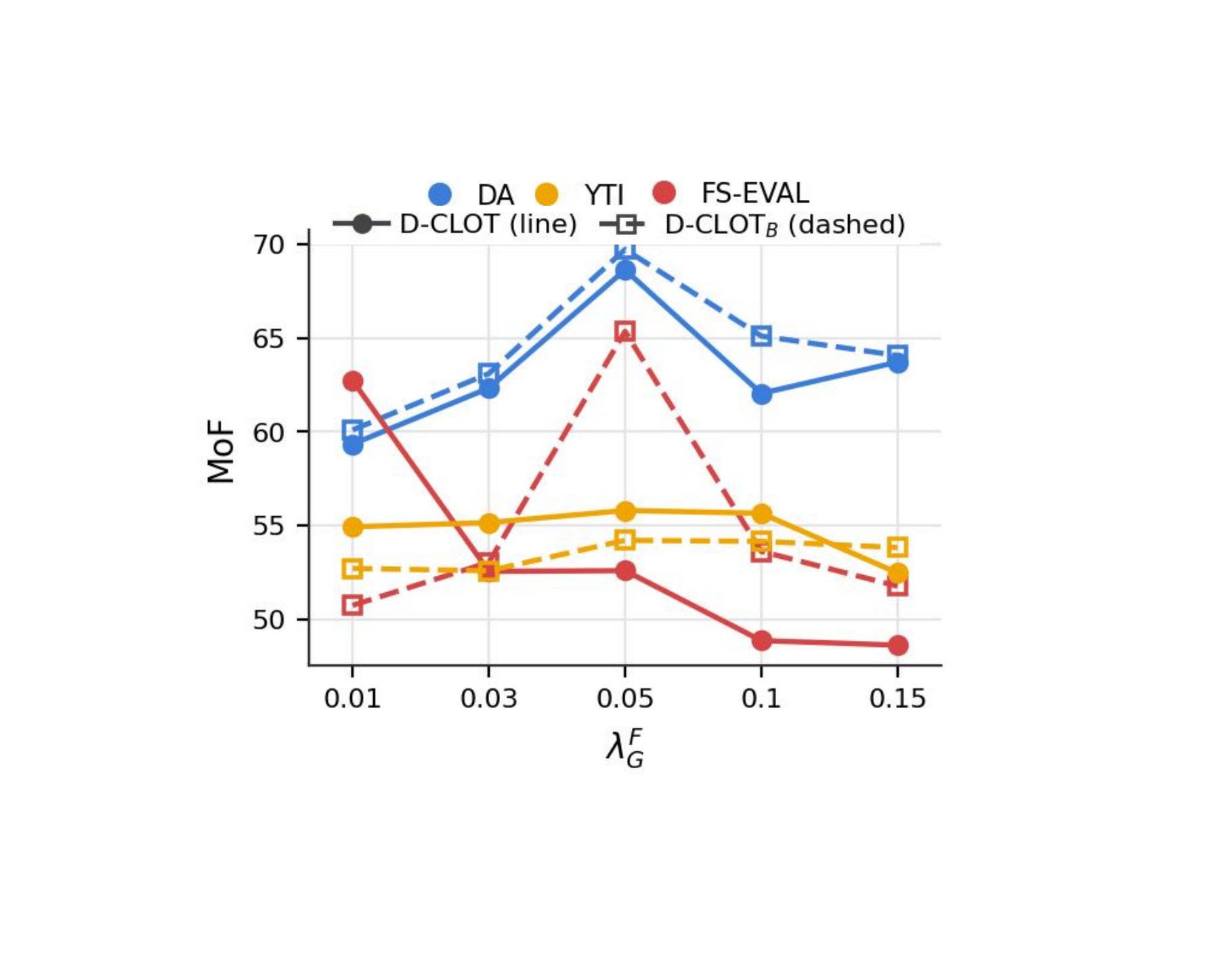}
    \vspace{-15mm}
    
    {\footnotesize (a) Frame graph weight $\lambda_G^F$}
    \vspace{-2em}
\end{minipage}
\begin{minipage}[t]{0.47\columnwidth}
    \centering
    \hspace*{-0.25\linewidth}
    \includegraphics[width=1.65\linewidth]{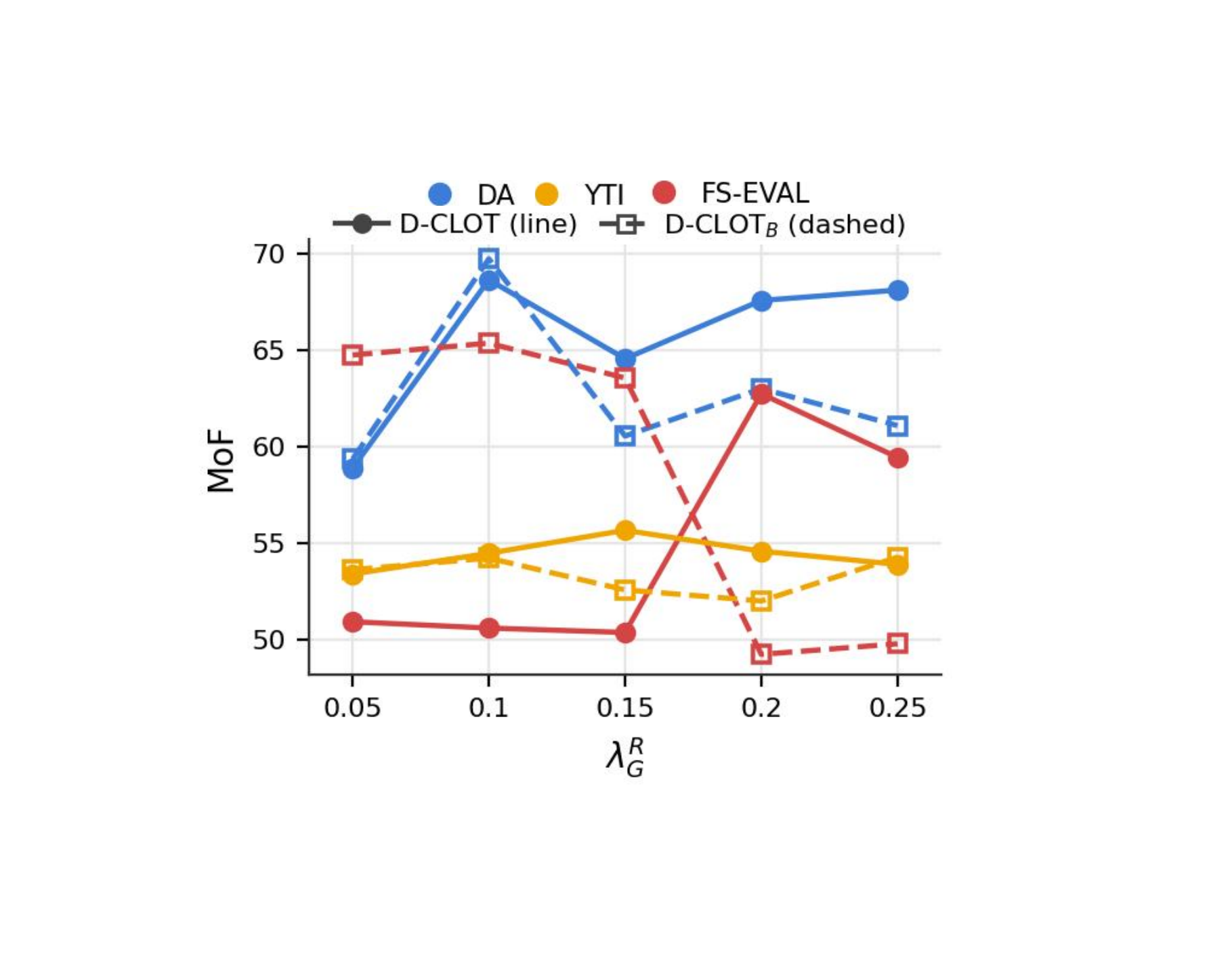}
    \vspace{-15mm}
    
    {\footnotesize (b) Refine graph weight $\lambda_G^R$}
    \vspace{-2em}
\end{minipage}
\vspace{-3em}

\begin{minipage}[t]{0.47\columnwidth}
    \centering
    \hspace*{-0.25\linewidth}
    \includegraphics[width=1.65\linewidth]{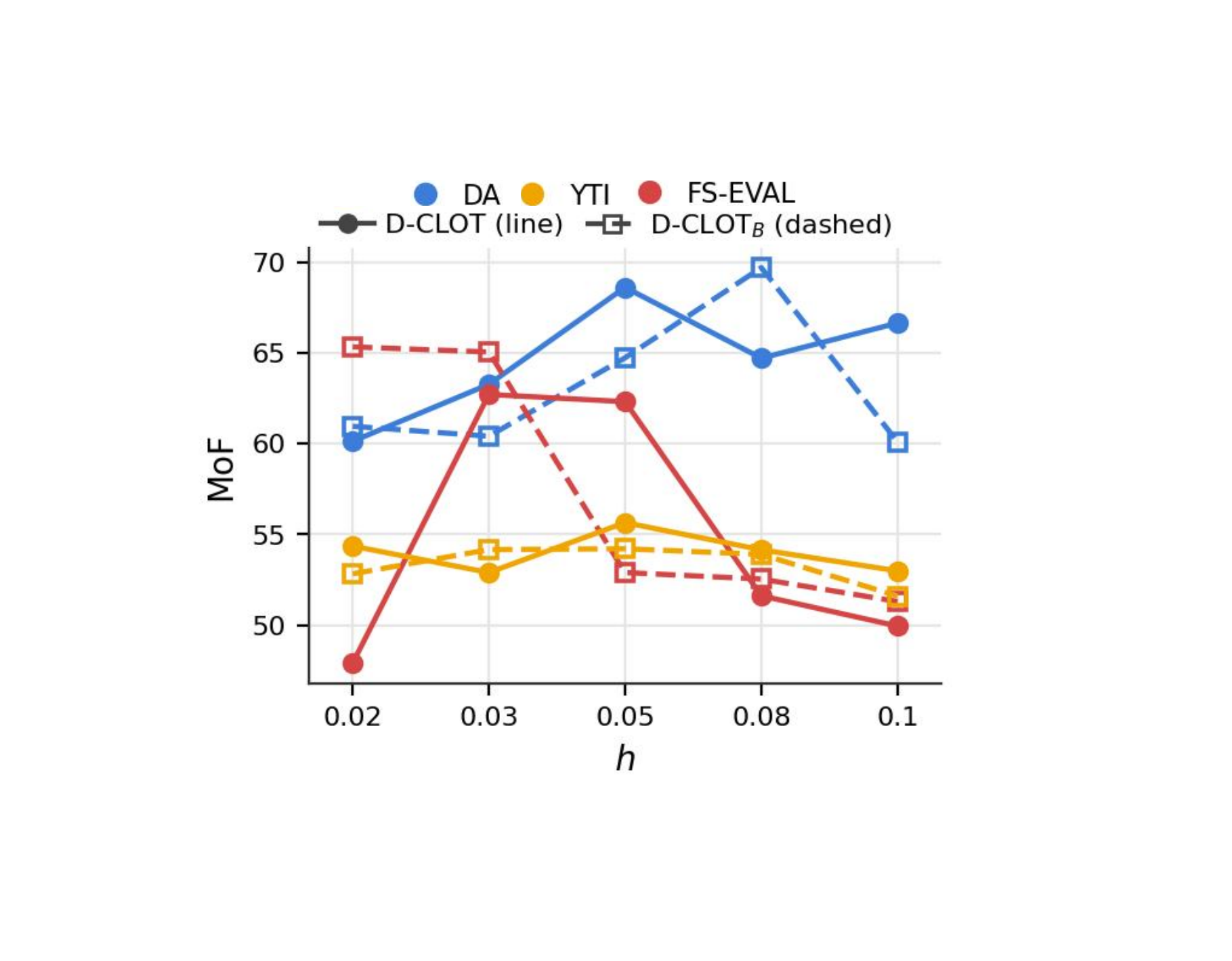}
    \vspace{-15mm}
    
    {\footnotesize (c) Graph bandwidth $h$}
    \vspace{-2em}
\end{minipage}
\hspace{-0.025\columnwidth}
\begin{minipage}[t]{0.47\columnwidth}
    \centering
    \hspace*{-0.25\linewidth}
    \includegraphics[width=1.65\linewidth]{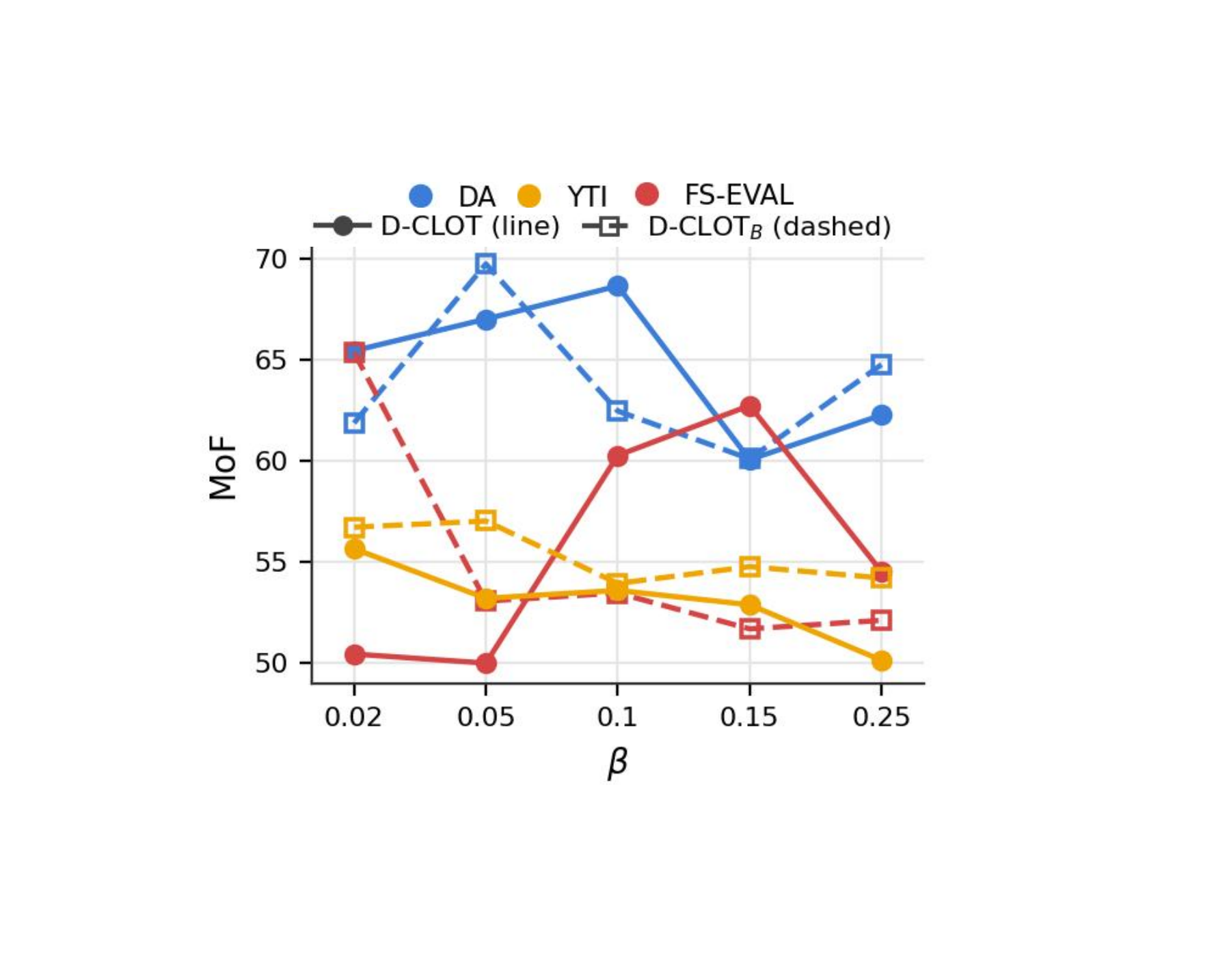}
    \vspace{-15mm}
    
    {\footnotesize (d) Blend factor $\beta$}
    \vspace{-2em}
\end{minipage}
\vspace{-3em}

\begin{minipage}[t]{0.47\columnwidth}
    \centering
    \hspace*{-0.25\linewidth}
    \includegraphics[width=1.5\linewidth]{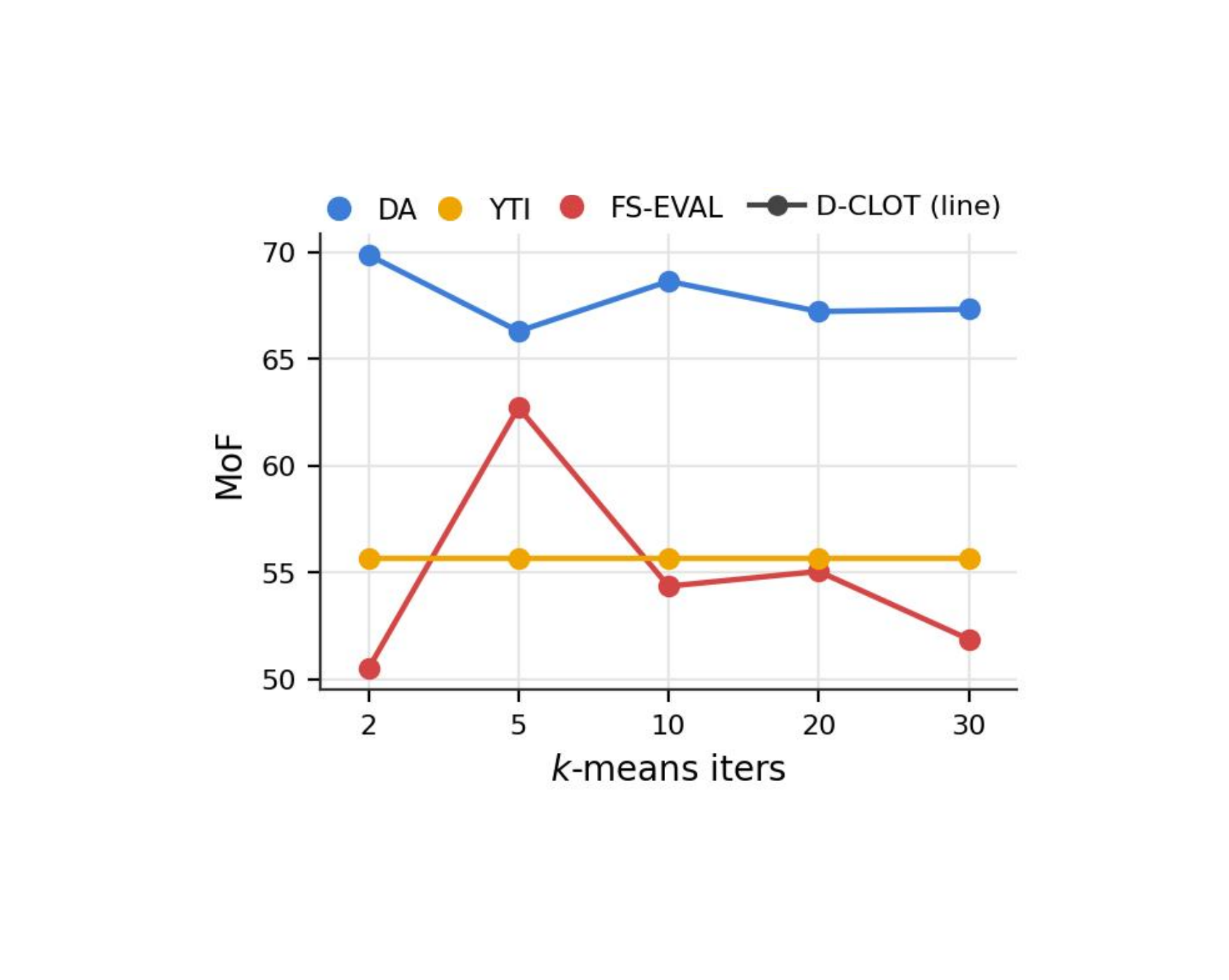}
    \vspace{-15mm}
    
    {\footnotesize (e) $k$-means refinement}
    \vspace{-2em}
\end{minipage}
\hspace{-0.025\columnwidth}
\begin{minipage}[t]{0.47\columnwidth}
    \centering
    \hspace*{-0.25\linewidth}
    \includegraphics[width=1.5\linewidth]{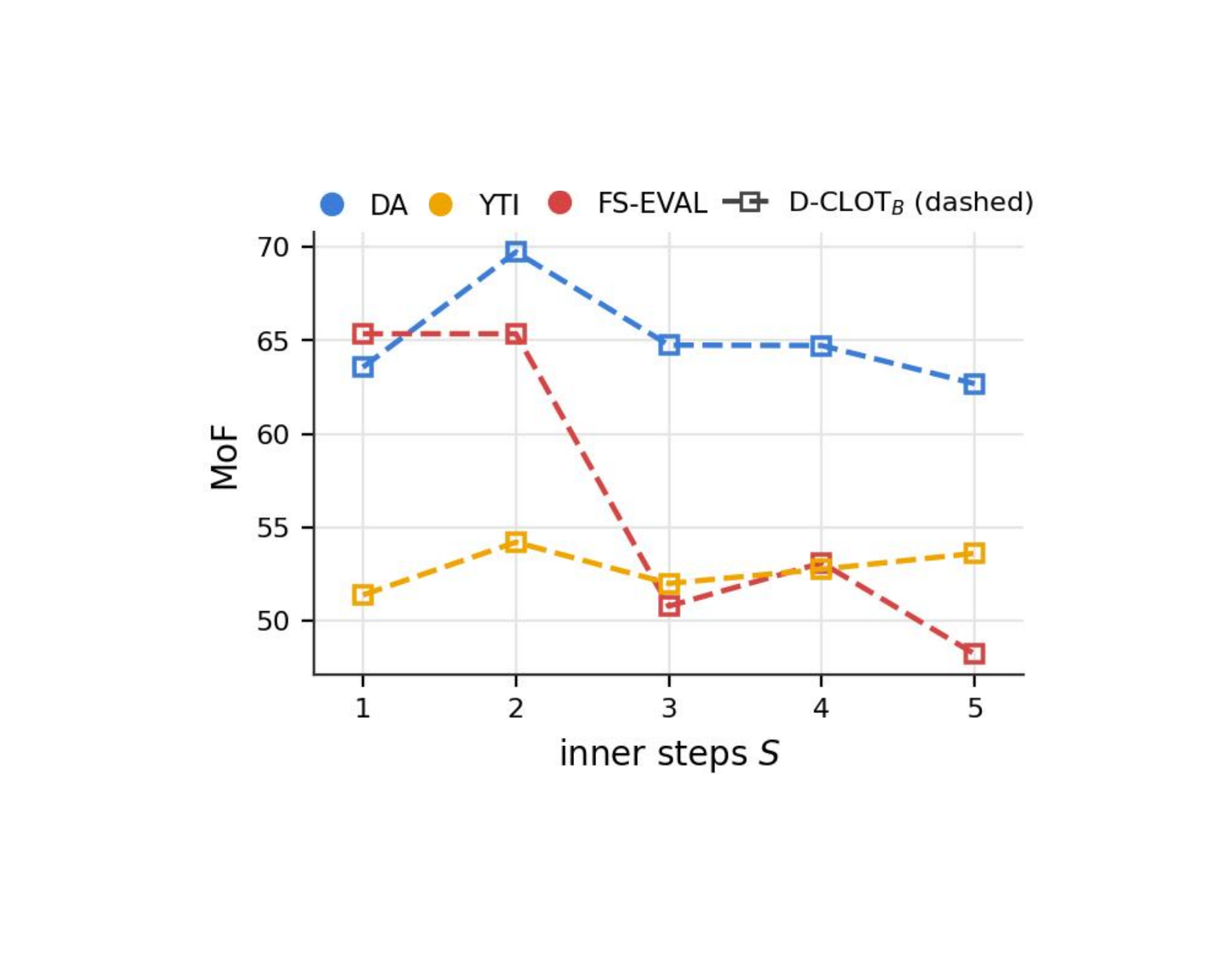}
    \vspace{-15mm}
    
    {\footnotesize (f) Barycentric refinement}
    \vspace{-2em}
\end{minipage}

\vspace{-1mm}
\caption{
Sensitivity analysis on mof. Each panel varies one hyperparameter while keeping the remaining configuration fixed. 
Colours denote datasets and line styles denote CLOT$^{2}$ and CLOT$^{2}_{B}$.
}
\label{fig:sensitivity_mof}
\end{figure}

\section{Computacional cost and practicality}
\label{sec:computational}
D-CLOT and D-CLOT$_{\mathrm{B}}$ retain the same 92.5K trainable parameters as CLOT. Their improvements therefore do not rely on additional model capacity. Table~\ref{tab:bf_complexity} reports training time, inference latency, and peak GPU memory on the Breakfast \texttt{coffee} activity. All methods use the same encoder, OT configuration, batch size, and hardware. D-CLOT adds 2.06\,s per training epoch due to the periodic $K$-means prototype update. This operation is used only during training, and its inference latency remains comparable to CLOT. D-CLOT$_{\mathrm{B}}$ adds 0.47\,s per epoch. Its barycentric re-estimation costs only 26\,ms every ten epochs. At inference, the inner refinement increases latency by 8.2\,ms per video. Peak memory remains below 82\,MB for all variants. Overall, the proposed closed loops introduce limited computational overhead and require no additional trainable parameters. D-CLOT preserves the inference cost of CLOT, while D-CLOT$_{\mathrm{B}}$ provides a stronger assignment-aware refinement at a moderate additional cost.

\begin{table}[t]
\centering 
\caption{Training cost of the CLOT family on the Breakfast \texttt{coffee} activity ($K{=}7$, 167 training videos, $B{=}2$, and $T{=}256$ sampled frames per video). Latency is averaged over 60 iterations after 15 warm-up iterations on a single RTX~4090. The refinement column reports its cost and update period.} \label{tab:bf_complexity} 
    \begin{threeparttable}
        \resizebox{\linewidth}{!}{%
        \begin{tabular}{lrrrrrr} 
            \toprule 
            Method & Params & GFLOPs & Iter. (ms) & Memory (MB) & Refinement & Epoch (s) \\ 
            \midrule
            CLOT & 92.5 & 0.036 & $61.7\pm0.8$ & 70.9 & -- & 5.02 \\ 
            D-CLOT & 92.5 & 0.077 & $63.8\pm0.6$ & 75.5 & 1,915 ms / epoch & 7.08 \\ 
            D-CLOT$_{\mathrm{B}}$ & 92.5 & 0.077 & $73.2\pm1.1$ & 81.2 & 26 ms / 10 epochs & 5.49 \\ 
            \bottomrule 
        \end{tabular}} 
    \end{threeparttable}
\end{table}
\section{More qualitative results}
\label{sec:more_examples}
\begin{figure}[t]
    \centering
   \begin{minipage}[t]{ \columnwidth}
    \centering 
    \includegraphics[width=1\linewidth,trim={0 50 0 50},clip]{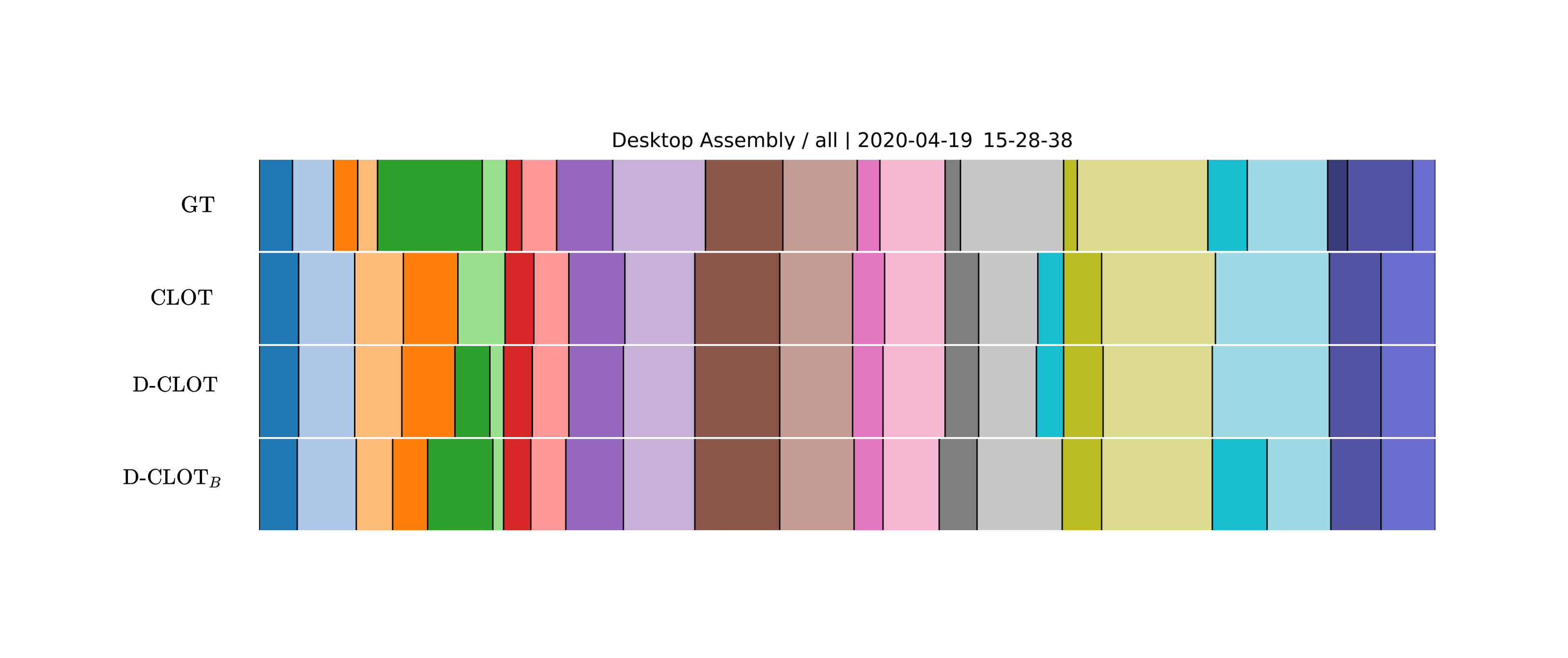}  
   
\end{minipage} 
\begin{minipage}[t]{ \columnwidth}
    \centering
    \vspace{-1.5em}
    \includegraphics[width=1\linewidth,,trim={0 100 0 50},clip]{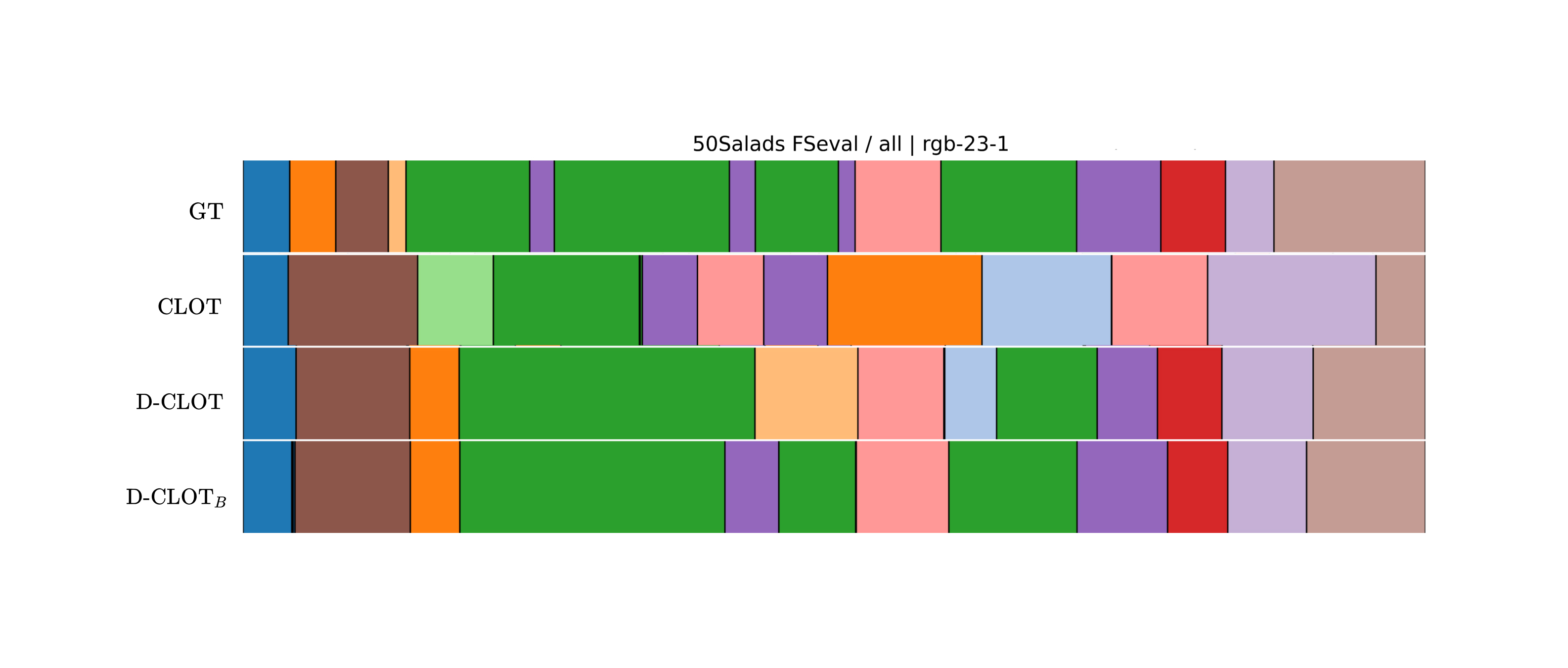} 
 
\end{minipage} 
    \par 
    
\caption{Example segmentation from DA\cite{Kumar22} (top) and FS (Eval)\cite{50salads} (bottom). We report the ground-truth (GT), the results of CLOT, and the result in ours two variants, D-CLOT and D-CLOT$_{B}$.}

\vspace*{-4mm}
 \label{fig:qualitative_segmentation_SUPP}
\end{figure}

\begin{figure}[t]
    \centering 
    
   \begin{minipage}[t]{ \columnwidth}
    \centering 
    \includegraphics[width=0.95\linewidth,trim={0 50 0 50},clip]{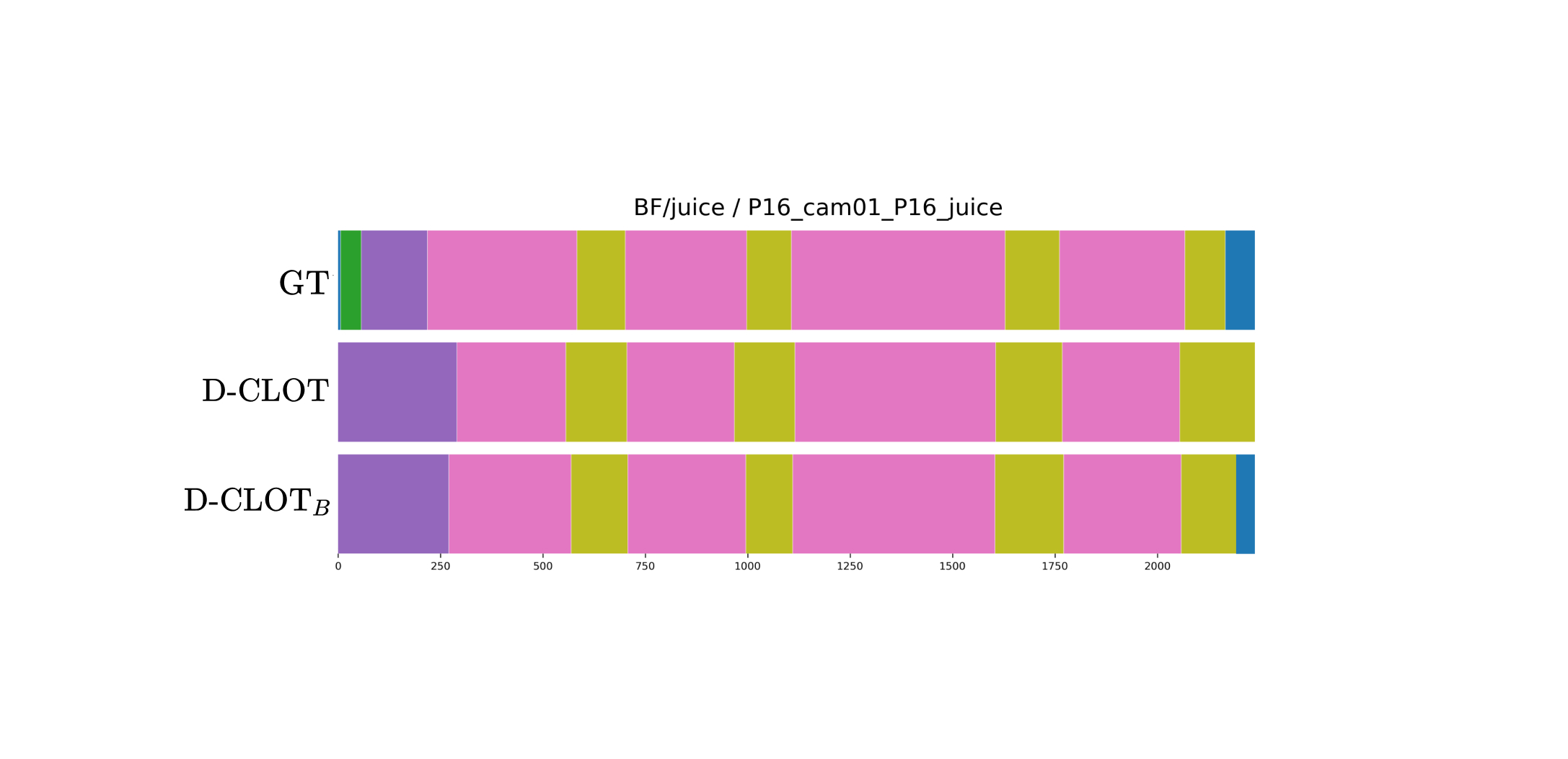}
    \vspace{-2em}
     
\end{minipage}  
 
   \begin{minipage}[t]{ \columnwidth}
    \centering
    \includegraphics[width=0.95\linewidth,trim={0 50 0 50},clip]{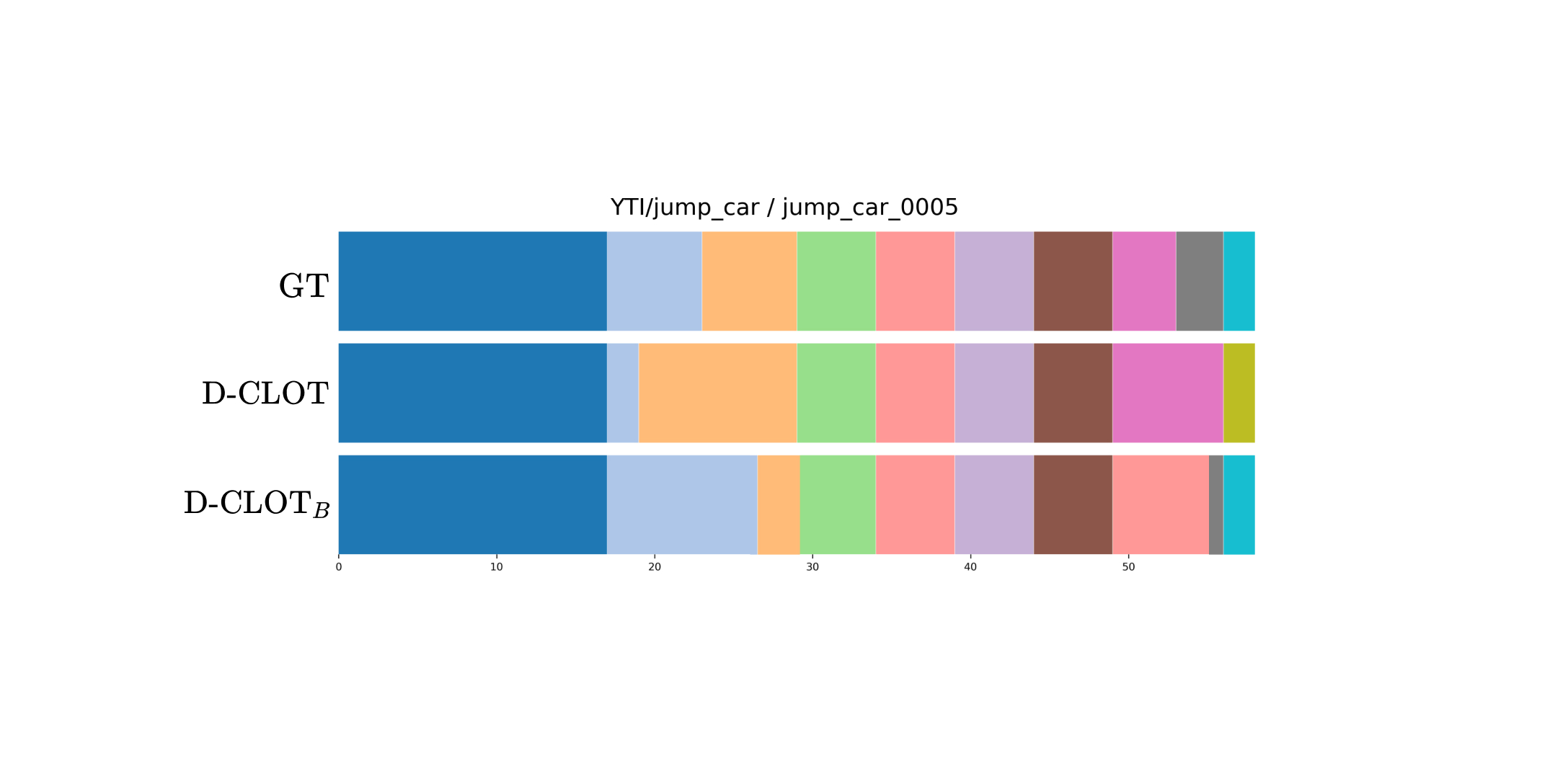}
    \vspace{-2em}

\end{minipage} 
\caption{Example segmentation from Breakfast\cite{breakfast} (top) and YTI\cite{ytii} (bottom), We report the ground-truth (GT), the results of CLOT, and the result in ours two variants, D-CLOT and D-CLOT$_{B}$.}

\vspace*{-4mm}
 \label{fig:qualitative_segmentation_2}
\end{figure}

Figure~\ref{fig:qualitative_segmentation_SUPP} provides additional qualitative comparisons across a broader set of videos. The results reinforce the trends observed in the main paper: D-CLOT and D-CLOT$_{B}$ better preserve temporal continuity, recover short and ambiguous actions, and reduce both over-segmentation and boundary drift relative to CLOT. In particular, the barycentric refinement of D-CLOT$_{B}$ yields more coherent action segments in challenging transition regions. The same behaviour appears on BF and YTI in Figure~\ref{fig:qualitative_segmentation_2}: both variants track the long dominant actions well, but D-CLOT tends to drop the short terminal segments (the trailing blue on BF, the closing teal on YTI), whereas D-CLOT$_{B}$ recovers them and preserves the GT ordering more faithfully. This is consistent with the quantitative trend: D-CLOT$_{B}$ preserves exactly the short, imbalanced actions that drive F1/mIoU while barely affecting the long-action-dominated MoF.

\vfill

\end{document}